\documentclass[11pt]{article}

\usepackage[preprint]{acl}

\usepackage{times}
\usepackage{latexsym}
\usepackage[T1]{fontenc}

\usepackage[utf8]{inputenc}

\usepackage{microtype}

\usepackage{inconsolata}

\usepackage{graphicx}
\usepackage{booktabs}

\usepackage{subcaption}
\title{(How) Do MLLMs Report Bistable Images Like Humans?\thanks{Accepted to EMNLP 2026 (Main Conference).}}

\author{
 \textbf{Ryota Takatsuki\textsuperscript{1,2}}\thanks{Work partly done at NII LLMC.},
 \textbf{Tomoki Doi\textsuperscript{3,4}},
 \textbf{Amane Watahiki\textsuperscript{3,4}},
 \textbf{Anil K. Seth\textsuperscript{1,5}},
 \textbf{Hitomi Yanaka\textsuperscript{3,4,6}}
\\
 \textsuperscript{1}Sussex Centre for Consciousness Science, University of Sussex,
 \textsuperscript{2}AI Alignment Network,
\\
 \textsuperscript{3}The University of Tokyo,
 \textsuperscript{4}RIKEN,
 \textsuperscript{6}Tohoku University,
\\
 \textsuperscript{5}Program on Brain, Mind, and Consciousness, Canadian Institute for Advanced Research
\\
 \small{
   \textbf{Correspondence:} \href{mailto:R.Takatsuki@sussex.ac.uk}{R.Takatsuki@sussex.ac.uk}
 }
}

\begin{document}
\maketitle

\begin{abstract}
Bistable images such as the duck--rabbit are classic stimuli in which one image supports multiple mutually incompatible interpretations, typically reported one at a time in humans. We ask whether multimodal large language models (MLLMs) show similar report behavior and what internal computations support it. Using the LLaVA family, we study two tractable dimensions: \emph{modulability}, whether reports can be biased by bottom-up visual cues and top-down linguistic priors, and \emph{exclusivity}, whether responses commit to a single interpretation. We test both on the canonical duck--rabbit and on synthetic Visual Anagrams to mitigate memorization confounds. Behaviorally, both visual and linguistic manipulations systematically shift reports in human-consistent ways, while responses remain predominantly exclusive. Mechanistically, these effects arise from competing image-token representations, distinct pathways for bottom-up and top-down modulation, and a link between exclusive reporting and object-count encoding.\footnote{Code and data are available at: \url{https://github.com/rtakatsky/mllm-bistable-images}.}
\end{abstract}

\section{Introduction}

\begin{figure}[!t]
    \centering
    \includegraphics[width=0.9\linewidth]{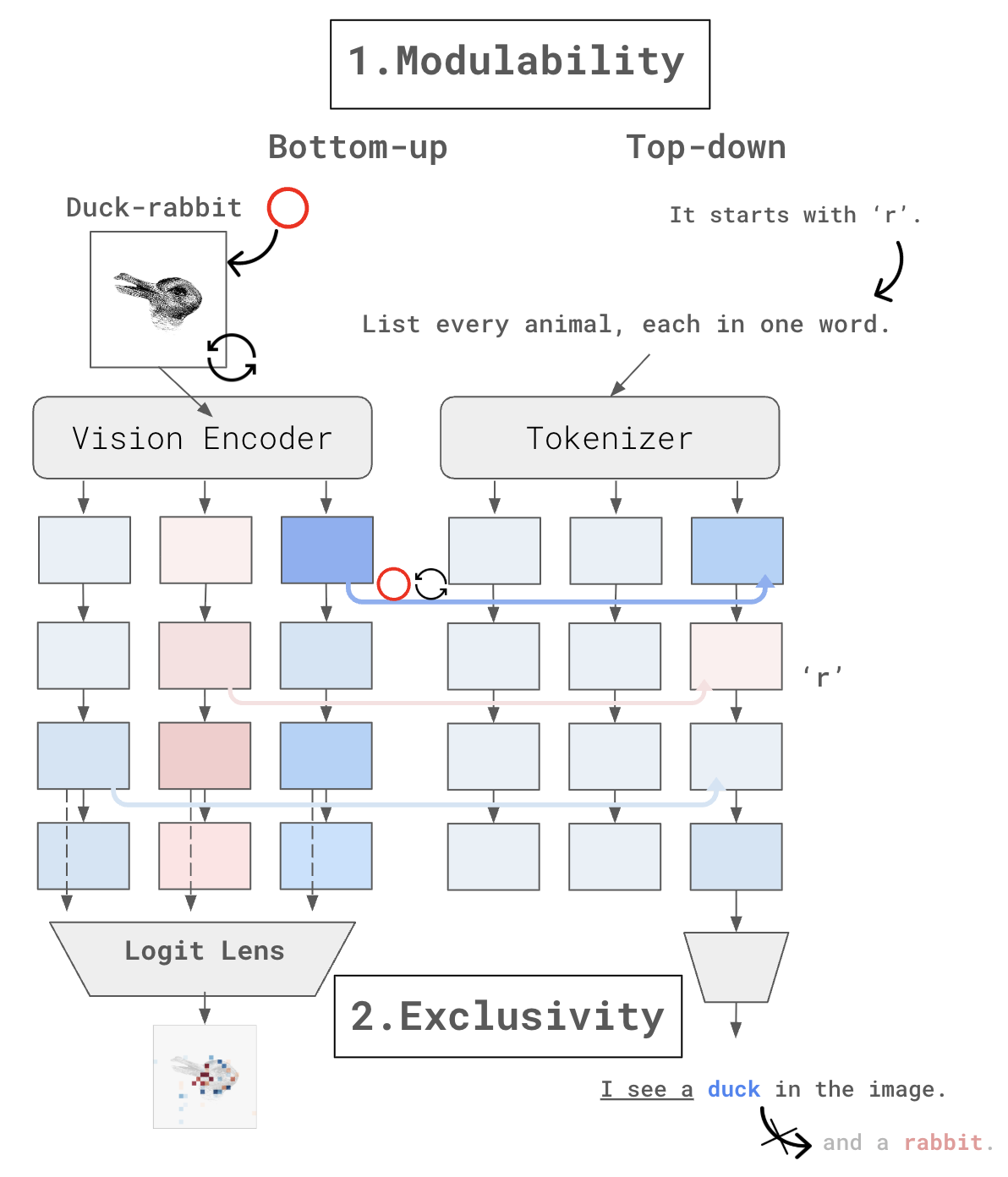}
    \caption{Overview of our framework. We test whether MLLM reports on the duck--rabbit show \emph{modulability} and \emph{exclusivity}, and then analyze how these behaviors are implemented along the vision-to-language pipeline.}
    \label{fig:main_fig}
\end{figure}

Multimodal large language models (MLLMs) have achieved strong performance across vision--language tasks \cite{Yin2024-qf}, but how they handle ambiguity in those inputs remains poorly understood. Following Simon's distinction between AI as engineering and AI as cognitive simulation \cite{Simon1983-fd}, this question matters both for building reliable systems and for understanding how ambiguous visual evidence can be transformed into overt report. Even if MLLMs and biological brains differ substantially at the implementational level, examining whether MLLMs generalize to such underdetermined cases in human-consistent ways, and how they do so, may reveal concrete computational mechanisms that emerge through learning to perform broadly analogous vision--language tasks, providing a comparative case for probing which aspects of such processing generalize across systems and which are system-specific \cite{Marr1976-tc,Vilas2024-eu}.

Bistable images such as the duck--rabbit \cite{Jastrow1899-fg} provide a particularly clean probe: the same static image supports multiple mutually incompatible interpretations, and in humans the reported interpretation can be biased by both visual and contextual factors \cite{Lupyan2013-hj,Intaite2013-cv}. For the present comparison, we focus on two tractable dimensions: \emph{modulability}, whether reported interpretations can be systematically biased by visual or linguistic manipulations, and \emph{exclusivity}, whether the system commits to a single interpretation in its report. We adopt these axes because they can be operationalized cleanly in end-to-end vision-to-language behavior without requiring sustained perception over time (see Limitations for other dimensions of human bistable perception).

Figure~\ref{fig:main_fig} summarizes our overall framework. Behaviorally, we test whether MLLMs show human-like modulability under bottom-up visual manipulations and top-down linguistic cues, while maintaining predominantly exclusive reports. Mechanistically, we ask how these report patterns are realized inside the model, focusing on competition between image-token representations, the distinct pathways underlying bottom-up and top-down modulation, and the relation between exclusivity and object-count encoding.

Most closely related to the present study is \citet{Panagopoulou2024-xx}, who benchmark multiple MLLMs on diverse bistable figures and emphasize ways in which current models diverge from humans. Our work asks a complementary question: whether there are tractable dimensions on which human-like report behavior does emerge, and what internal mechanisms support it. Because famous bistable images may also be memorized by web-scale models \cite{Ullman2024-pd,Shinozaki2025-qv}, we test not only the canonical duck--rabbit but also synthetic Visual Anagram stimuli \cite{Geng2024-be}. Across five LLaVA variants, we show that human-like bistable report behavior can emerge in identifiable computational forms, even in systems whose underlying architecture differs substantially from the brain.

\section{Related Work}

\subsection{Bistable perception in humans}

Bistable perception arises when a constant stimulus supports multiple mutually incompatible interpretations \citep{Leopold1999-ae,Leopold2002-dx,Brascamp2018-gq}. The duck--rabbit is a canonical example with many published variants \citep{Jastrow1899-fg,Brugger1999-rh,Kihlstrom2004-su}; relatively small changes in image orientation, or in which local features are fixated, can shift which interpretation is reported \citep{Moreno-Sanchez2016-og,Hsu2025-sj}. Humans also typically report one interpretation at a time, even though the same image supports multiple alternatives. This makes bistable images a useful setting for comparing human and model report behavior under ambiguity.

The two properties we focus on, \emph{modulability} and \emph{exclusivity}, form a natural decomposition for ambiguous figures because, unlike binocular rivalry, they are more readily influenced by top-down factors such as attention, context, and prior knowledge \citep{Meng2004-zh,Scocchia2014-el}. Consistent with this, bistable reports are shaped by bottom-up factors such as image rotation and eye movements \citep{Moreno-Sanchez2016-og,Hsu2025-sj}, as well as by top-down factors such as language, expectation, and audiovisual context \citep{Brugger1993-go,Lupyan2017-jy,Lupyan2013-hj,Hsiao2012-ww}; these influences can make partly independent, approximately additive contributions to perceptual selection and report \citep{Intaite2013-cv}.

\subsection{Bistable perception and relevant phenomena in MLLMs}
A growing body of work evaluates vision--language systems on visual illusions as probes of robustness and human-likeness. The work most closely related to ours is the benchmark of bistable images introduced by \citet{Panagopoulou2024-xx}. They evaluate 12 MLLMs on a diverse set of ambiguous figures, including the duck--rabbit, and show that models typically favor a single interpretation, that most image-level manipulations have limited effect on that choice (with rotation as a notable exception), and that relatively small prompt variations can substantially change outputs. Their results emphasize ways in which current MLLMs differ from humans in resolving bistable images. Our study complements this breadth-first benchmark by focusing on a single canonical stimulus and asking not only which interpretation is reported, but also what determines the report behaviorally and mechanistically.

Beyond bistable figures, prior work has also tested MLLMs on classical visual illusions, including size, geometry, brightness, and color illusions \citep{Zhang2023-grounding,Shahgir2024-wf,Mao2025-uc}. Across these settings, model behavior depends strongly on architecture, prompting, and task formulation, and larger models sometimes show more human-like illusion-consistent judgments. This broader literature motivates treating illusions as a useful testbed for how MLLMs transform visual input into language under conditions where appearance systematically departs from physical reality. A central complication, however, is that internet-trained models may "know" about famous illusions without necessarily processing them in a stimulus-driven way: recent work shows that MLLMs can report illusion-consistent effects even for fake or trap variants that only resemble well-known illusion classes \citep{Ullman2024-pd,Shinozaki2025-qv,Zhang2025-illusionbenchp}.

To mitigate this confound, we employ recent work that has used optimization and generative methods to construct novel illusion-like images \citep{Gomez-Villa2022-wr,Gomez-Villa2024-si}. Particularly relevant here are \emph{Visual Anagrams} \citep{Geng2024-be}, which produce transformation-dependent ambiguity under rotation or reflection while being less likely to have been memorized as canonical illusion exemplars. Related benchmarks use pareidolia-like stylized ambiguous images to control for memorization \citep{Hemmat2024-rg,Anvekar2026-qp}, but such stimuli are, though related, distinct from bistable images studied here \citep{Lhotka2022-an}.

Two additional lines of work are especially relevant for our framing of modulation: prompt-based steering in shape--texture conflict settings \citep{Gavrikov2024-pc} and visual-prompt interventions such as circles or arrows \citep{Shtedritski2023-ai,Cai2024-on,Zhuang2024-pw}. Our work brings these strands together in a single bistable stimulus, comparing top-down prompt priors with bottom-up visual cueing under conditions designed to reduce memorization effects.

\subsection{Interpreting visual-to-text processing in MLLMs}

One prominent approach to understanding the inner workings of MLLMs is information-flow analysis via attention ablations or ``attention knockout'' \citep{Geva2023-dr}, where selected attention weights are intervened on to test causal contributions. Such methods can reveal how image-conditioned information propagates into generated tokens, and have shown that later-layer attention from generated tokens back to relevant image regions can be critical for grounded visual descriptions \citep{Kaduri2025-yw,Zhang2024-crossmodal,Jiang2024-devils}. This provides a natural tool for our setting, where we ask whether bottom-up and top-down manipulations alter which image tokens causally influence the final report.

A complementary line of work applies probing and logit-lens-style analyses to image token representations \citep{nostalgebraist2020-logitlens,Neo2024-ep}. These methods make it possible to localize visual information across layers, which can be used to mitigate visually ungrounded outputs \citep{Jiang2024-mitigate,Che2025-pu}. 

Related work has examined how MLLMs handle conflicts between information sources, such as visual input versus learned world knowledge utilizing multimodal counterfactuals \citep{Golovanevsky2025-am,Ortu2025-iq} or image objects versus contradictory in-image captions \citep{Hua2025-tn}. Our setting is different: we study \emph{intrinsic} ambiguity within a single image, where one visual input supports multiple incompatible interpretations, rather than competition between vision and external or in-image text information.

\section{Behavioral Analyses}

\begin{figure*}[t]
    \centering
    \includegraphics[width=\textwidth]{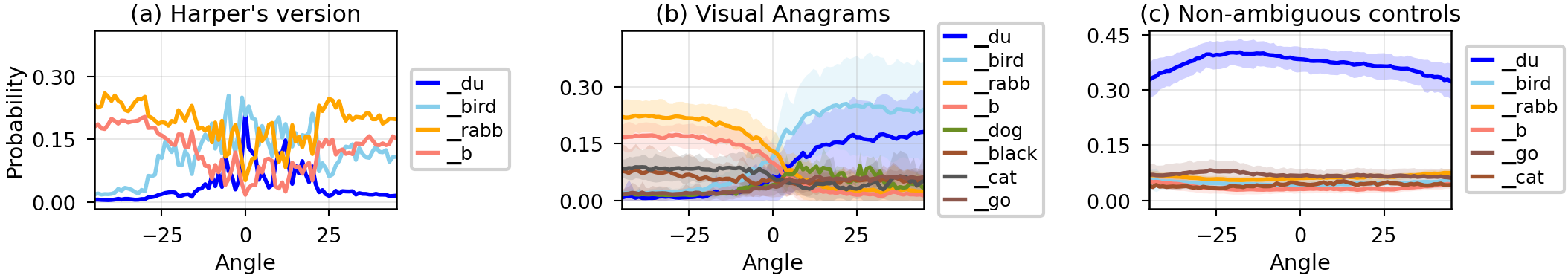}
    {\phantomsubcaption\label{fig:behavioral_rotation_harper}}
    {\phantomsubcaption\label{fig:behavioral_rotation_va}}
    {\phantomsubcaption\label{fig:behavioral_rotation_split}}
    \caption{Bottom-up modulation by rotation in \textit{LLaVA-1.5-7B} under the default query and prefix. Small changes in image orientation shift the relative preference for duck- and rabbit-related outputs for Harper's version (a) and the generated ambiguous stimuli (b), demonstrating modulability. The two-animal controls (c) show no such reversal: the duck-related first token remains dominant across the same rotation range. In (b), (c), and all aggregated plots in this paper, lines indicate the mean across valid input images and shading indicates $\pm 1$ standard deviation.}
    \label{fig:behavioral_rotation}
\end{figure*}

\subsection{General Setup}

\paragraph{Visual input.}
To facilitate comparison with human perceptual studies, we use the widely adopted Harper's version of the duck--rabbit \cite{Kihlstrom2004-su}. Although none of the tested models explicitly recognized the duck--rabbit image as a visual illusion via a general description query, we also construct 50 duck--rabbit-like images using \textit{Visual Anagrams} \citep{Geng2024-be}, each designed to yield different views under 90-degree rotation, via an automated pipeline with an external VLM-judge validity filter. For each ambiguous stimulus, we also construct matched non-ambiguous controls: a style-matched \emph{two-animal control} showing a duck and a rabbit side by side, and the corresponding duck-only and rabbit-only \emph{single-animal controls}. Details of stimulus construction and preprocessing are provided in Appendix~\ref{sec:appendix_setup}.

\paragraph{Models.}
We study five LLaVA-family models: \textit{LLaVA-1.5-7B}, \textit{LLaVA-1.5-13B} \citep{Liu2024-llava15}, \textit{LLaVA-v1.6-Vicuna-7B}, \textit{LLaVA-v1.6-Mistral-7B}, and \textit{Llama3-LLaVA-Next-8B} \citep{Liu2024-llavanext}. These models build on the LLaVA framework \citep{Liu2023-llava}, which combines a CLIP vision encoder \citep{Radford2021-nt} with a pretrained language model through a learned vision--language projector. Because they share the same visual backbone (CLIP-ViT-L), variation in early visual encoding is reduced, allowing us to focus on downstream processing in the language model. Though we use the LLaVA family for the main depth-first mechanistic analyses, in order to test whether the behavioral patterns are specific to this architecture family, we additionally run focused behavioral checks on four non-LLaVA models---\textit{Qwen2-VL-7B-Instruct} \citep{Wang2024-qwen2vl}, \textit{SmolVLM-Instruct} \citep{Marafioti2025-smolvlm}, \textit{IDEFICS2-8B} \citep{Laurencon2024-idefics2}, and \textit{InstructBLIP-Vicuna-7B} \citep{Dai2023-instructblip}.

For space reasons, the main text presents detailed results for \textit{LLaVA-1.5-7B} as a representative model, with results for the other LLaVA-family models provided in Appendix~\ref{sec:appendix_llava_all_models} and for non-LLaVA models in Appendix~\ref{sec:appendix_non_llava}.

\paragraph{Evaluation.}
Unless otherwise noted, we query the model with \texttt{"List every animal in the image, each in one word."} and use \texttt{"I see a"} as the output prefix. We collect responses in a single-turn setting. To quantify \emph{modulability}, we examine the next-token probability distribution following the prefix while manipulating either the image itself (bottom-up modulation) or linguistic cues in the query (top-down modulation). To quantify \emph{exclusivity}, we use beam search. For single images we track the top beam continuations directly; for aggregates over many stimuli we classify each continuation with an LLM judge as naming one or multiple animals and report probability-mass shares (Appendix~\ref{sec:appendix_beam_classification}). We additionally use an explicit object-count query to probe whether exclusive reporting is associated with construing the ambiguous stimulus as a single object. Results with additional prompts are provided in Appendix~\ref{sec:appendix_across_prompts}.

\subsection{Modulability}
\label{sec:behavioral_modulability}

\subsubsection{Bottom-up Modulation}

We examine two forms of visually driven modulation: rotating the image, and adding a red circle to part of the image to increase bottom-up saliency at a specific location.

\paragraph{Rotation.}
We fix the text input and rotate the duck--rabbit image around its center in one-degree increments, placing it on a white background.
As Figure~\ref{fig:behavioral_rotation_harper} shows, for Harper's version at angle $=0^\circ$, all tested models assign higher probability to duck-related continuations (e.g., \texttt{"\_bird"} or \texttt{"\_du"}) than to rabbit-related continuations (e.g., \texttt{"\_rabb"} or \texttt{"\_b"}), indicating a dominant duck interpretation.\footnote{We use tokenizer-level strings such as \texttt{"\_bird"}, where the leading underscore denotes a preceding space.} Moreover, even a slight rotation is sufficient for the rabbit interpretation to become dominant, indicating that the region around angle $=0^\circ$ is highly unstable. These results are consistent with \citet{Panagopoulou2024-xx}.

For most Visual Anagram images (37--44 of 50, depending on the model; Appendix~\ref{sec:appendix_boundary_fit}), the interpretation reverses at a particular angle, resembling the pattern reported in human participants by \citet{Moreno-Sanchez2016-og}. These valid stimuli are used in the aggregate Visual Anagram analyses later on. To visualize this tendency across images, we align each image by defining angle $=0^\circ$ as the angle at which a logistic psychometric function fitted to the normalized first-token probabilities of the bird and rabbit tokens crosses 50\% (Appendix~\ref{sec:appendix_boundary_fit}). By contrast, the matched two-animal controls show no comparable reversal under rotation (Figure~\ref{fig:behavioral_rotation_split}; Appendix~\ref{sec:appendix_control_images}).

\paragraph{Red-circle cueing.}
To compare model behavior with human eye-movement results \cite{Hsu2025-sj}, we add a red circle to Harper's version and sweep it across the image (experimental details in Appendix~\ref{sec:appendix_red_circle}).

\begin{figure}
    \centering
    \includegraphics[width=\linewidth]{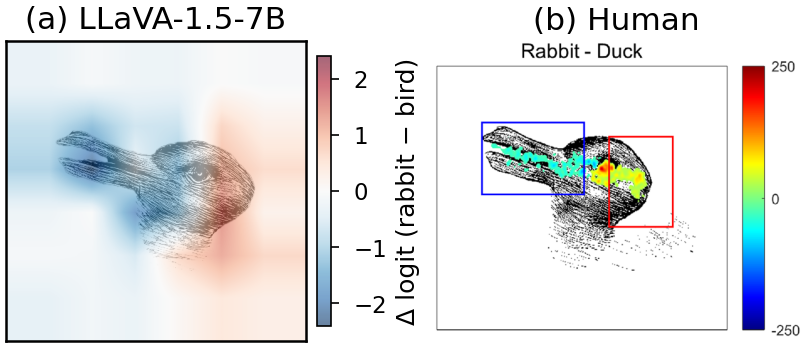}
    {\phantomsubcaption\label{fig:behavioral_red_circle_model}}
    {\phantomsubcaption\label{fig:behavioral_red_circle_human}}
    \caption{Bottom-up modulation by local visual cueing: rabbit-vs-duck difference maps for red-circle logit effects in \textit{LLaVA-1.5-7B} (a) and human fixation density (b), adapted from \citet{Hsu2025-sj}. Red-circle placement shifts the logit difference in regions qualitatively aligned with the human fixation-density pattern.}
    \label{fig:behavioral_red_circle}
\end{figure}

Rabbit-related output increases when the circle is placed near the rabbit's mouth, whereas duck-related output increases when it is placed near the duck's beak (Figure~\ref{fig:behavioral_red_circle_model}). This spatial pattern qualitatively aligns with the human fixation-density result in Figure~\ref{fig:behavioral_red_circle_human}, suggesting that competing interpretations are supported by similar local visual features.

\subsubsection{Top-down Modulation}
\label{sec:behavioral_topdown}

We next examine how strongly the reported interpretation can be biased by linguistic priors. Starting from the default query, we append either a prefix cue (e.g., \texttt{"It starts with 'r'."}) or a semantic cue (e.g., \texttt{"It's terrestrial."}).

\begin{figure}
    \centering
    \includegraphics[width=\linewidth]{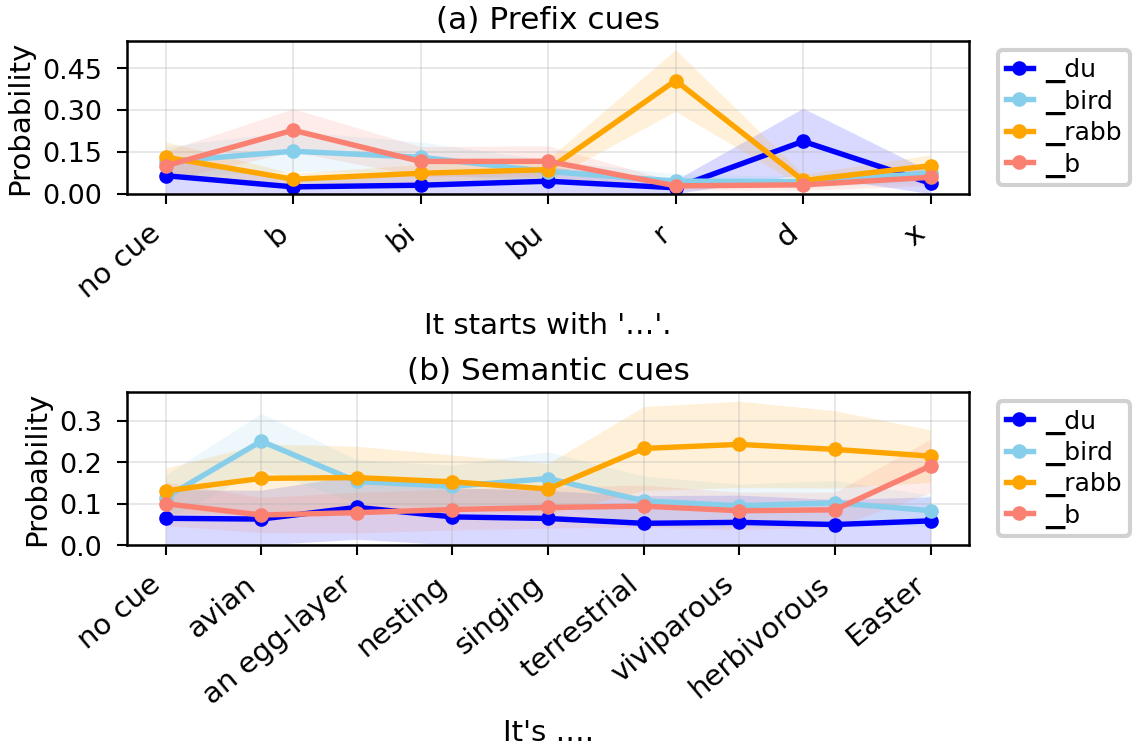}
    {\phantomsubcaption\label{fig:behavioral_top_down_prefix}}
    {\phantomsubcaption\label{fig:behavioral_top_down_semantic}}
    \caption{Top-down modulation by linguistic priors in \textit{LLaVA-1.5-7B}. Both prefix-based and semantic cues bias the reported interpretation toward the intended animal, with prefix cues producing the stronger effect.}
    \label{fig:behavioral_top_down}
\end{figure}

As shown in Figure~\ref{fig:behavioral_top_down}, both cue types successfully bias the output toward the intended interpretation: for example, \texttt{"It starts with 'r'"} substantially increases the output probability of the \texttt{"\_rabb"} token, and \texttt{"It's Easter."} strongly modulates the \texttt{"\_b"} token, consistent with human results \citep{Brugger1993-go}. This modulation is bounded by the image: applied to matched non-ambiguous controls that show only the uncued animal, the same cues leave the cued interpretation essentially absent from the output. Linguistic context therefore appears to bias competition between the interpretations an ambiguous image supports, rather than to induce the cued label whenever it is named. Additional analyses are reported in Appendices~\ref{sec:appendix_topdown_boundary_nonambig} and \ref{sec:appendix_topdown_boundary_negation}.

\subsection{Exclusivity}
\label{sec:behavioral_exclusivity}

We next ask whether MLLMs tend to report only one of the competing interpretations, rather than merely assigning different relative probabilities to both.

\paragraph{Exclusive reporting.}
For Harper's version, we include one interpretation directly in the prefix (e.g., \texttt{"I see a bird"} or \texttt{"I see a rabbit"}) and examine continuation behavior using beam search.

\begin{figure*}[t]
    \centering
    \includegraphics[width=\textwidth]{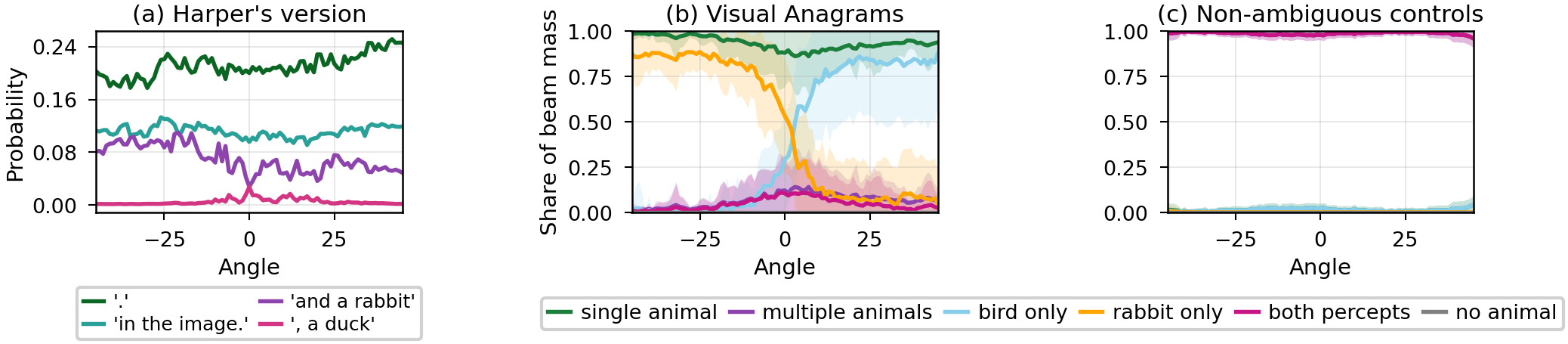}
    {\phantomsubcaption\label{fig:behavioral_rotation_beam_search_harper}}
    {\phantomsubcaption\label{fig:behavioral_rotation_beam_search_va}}
    {\phantomsubcaption\label{fig:behavioral_rotation_beam_search_split}}
    \caption{Exclusivity under beam search (\textit{LLaVA-1.5-7B}). (a) Harper's version with the bird interpretation forced into the prefix: single-interpretation continuations (\texttt{"."}, \texttt{" in the image."}) outweigh enumerating ones (\texttt{" and a rabbit"}); the rabbit-prefixed condition behaves alike (Figure~\ref{fig:appendix_beam_rabbit_all_models}). (b, c) Under the default prefix, each beam continuation is classified as naming a single or multiple animals (Appendix~\ref{sec:appendix_beam_classification}); curves are the mean $\pm$ SD across stimuli: ambiguous stimuli are reported almost exclusively as a single animal, the two-animal controls as multiple.}
    \label{fig:behavioral_rotation_beam_search}
\end{figure*}

As shown in Figure~\ref{fig:behavioral_rotation_beam_search_harper}, continuations corresponding to a single interpretation only, such as \texttt{"."} or \texttt{" in the image."}, are more probable than continuations such as \texttt{"\_and"} or \texttt{","}, which would naturally lead to listing both interpretations. This tendency weakens only slightly even when the prefix forces the interpretation opposite to the model's initial preference.

The same pattern extends to the synthetic stimuli. Aggregated over the Visual Anagrams (Figure~\ref{fig:behavioral_rotation_beam_search_va}), multiple-animal continuations receive only $\sim$10\% of beam mass near the ambiguity boundary and less elsewhere. By contrast, the matched two-animal controls are reported as multiple animals with $\geq$98\% of the beam mass (Figure~\ref{fig:behavioral_rotation_beam_search_split}). Thus, the model's tendency to report only one interpretation is specific to the ambiguous stimuli rather than a generic reluctance to enumerate multiple animals.

This result parallels the exclusivity of human ambiguous perception, in which only one interpretation enters awareness at a given moment. It also provides a quantitative complement to the qualitative observations of \citet{Panagopoulou2024-xx}.

\paragraph{Relation to object count.}
One possible basis for this exclusivity is that the two interpretations are construed as alternative identities of a single object rather than as two simultaneously present objects. To probe this possibility behaviorally, we ask \texttt{"How many objects are in the image? Answer with a number only."} All five LLaVA-family models answer \texttt{"1"} for Harper's version (0.88--0.99) and for the ambiguous Visual Anagrams at their boundary angles (0.88--0.99), whereas for the matched two-animal controls they answer \texttt{"2"} (0.87--0.99; Table~\ref{tab:appendix_object_count}). The object-count probe thus separates the ambiguous stimuli from the two-animal controls exactly where beam search separates exclusive from enumerating reports. This correspondence suggests a relationship between exclusive reporting and the model's object count, but does not by itself establish that the latter causes the former.

\paragraph{A figure--ground boundary condition.}
To examine whether the same pattern extends beyond object-category ambiguity, we additionally test the Raven--Bear figure--ground stimulus used in prior MLLM work \citep{Panagopoulou2024-xx}. Its reported interpretation remains modulable by both rotation and linguistic cues, but exclusivity weakens and, crucially, does so together with the object count: the three models with weaker single-object judgments (p(\texttt{"1"})~$=0.18$--$0.60$) allocate 43--51\% of the beam mass to naming both interpretations, whereas the two models with strong single-object judgments (0.84 and 0.94) remain fully exclusive (Table~\ref{tab:appendix_object_count}; Appendix~\ref{sec:appendix_figure_ground}). Figure--ground images can more readily support a two-entity construal than the duck--rabbit's single-region, two-identity ambiguity. This model-by-model covariation of exclusivity and object count provides a behavioral motivation for testing their causal relationship in the mechanistic analyses below.

\subsection{Robustness Across Prompts and Models}

Although the results are somewhat prompt-sensitive, as noted in prior work, the qualitative modulability and exclusivity patterns are observed across different prompts, including general description and zero-shot chain-of-thought (CoT) prompting \citep{Kojima2022-zs}, and across models including non-LLaVA architectures. An explicit two-animal hint can reduce exclusivity in a model-dependent manner, indicating that exclusive reporting is a default construal rather than a hard output constraint. Full results across prompts and models are provided in Appendices~\ref{sec:appendix_llava_all_models}, \ref{sec:appendix_non_llava}, and \ref{sec:appendix_across_prompts}.

\section{Mechanistic Analyses}

\begin{figure}[t]
    \centering
    \includegraphics[width=\linewidth]{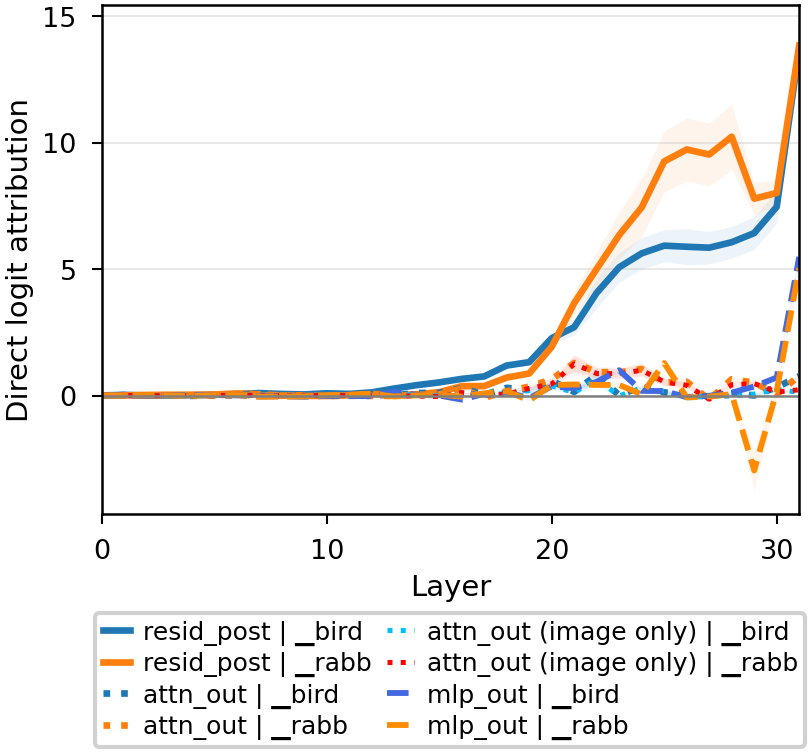}
    \caption{Layerwise direct logit attribution (DLA) for \texttt{"\_bird"} and \texttt{"\_rabb"} in \textit{LLaVA-1.5-7B}. Both interpretation logits rise in later layers, with image-token attention contributing most strongly in middle-to-late layers.}
    \label{fig:mechanistic_layerwise_dla}
\end{figure}

\begin{figure}
    \centering
    \includegraphics[width=\linewidth]{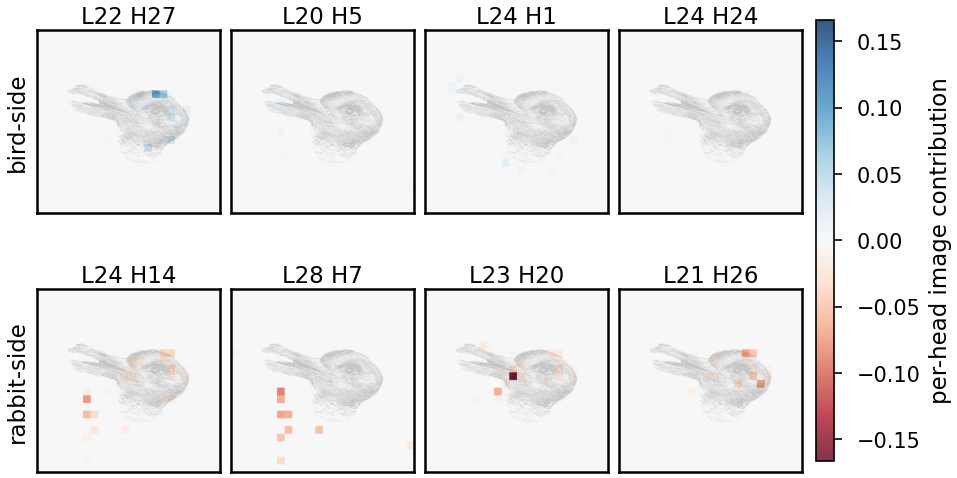}
    \caption{Attention maps of heads contributing most strongly to the bird--rabbit logit difference (blue: bird-side, red: rabbit-side), on the canonical duck--rabbit image. These heads attend to local regions supporting different interpretations.}
    \label{fig:mechanistic_topk_head_attn_maps}
\end{figure}

\begin{figure}
    \centering
    \includegraphics[width=0.5\linewidth]{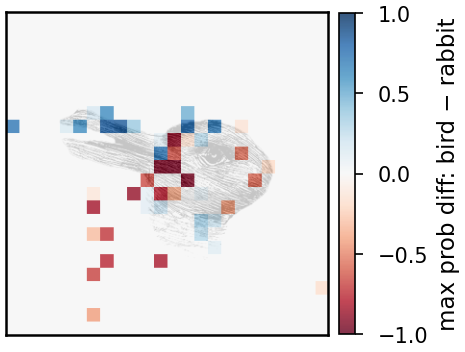}
    \caption{Example dominant patch map from image-token logit lens, on the same image (blue: bird-dominant, red: rabbit-dominant patches).}
    \label{fig:mechanistic_dominant_patch_map}
\end{figure}

In this section, we analyze \emph{modulability} and \emph{exclusivity} using methods from mechanistic interpretability \citep{Bereska2024-te}, combining observational analyses such as direct logit attribution (DLA) and attention-map analysis with interventional analyses such as activation patching. We focus on how competing interpretations are represented, selected, and read out on the language-model side of the MLLMs. 

We begin with a neutral case without explicit modulation, then analyze how bottom-up and top-down modulation alter the internal mechanism, and finally test the behavioral suggestion that exclusive reporting is related to object-count encoding. We show the results on \textit{LLaVA-1.5-7B}, and analyses involving Visual Anagram stimuli use the samples with a valid decision boundary for this model (44 of 50; Appendix~\ref{sec:appendix_boundary_fit}), evaluated at their boundary angles.

\subsection{Neutral Setup}

Figure~\ref{fig:mechanistic_layerwise_dla} shows DLA of the attention output (\texttt{attn\_out}), the multilayer perceptron output (\texttt{mlp\_out}), and the Transformer block output (\texttt{resid\_post}) at each layer for the output tokens \texttt{"\_bird"} and \texttt{"\_rabb"}. Following prior work on multimodal information flow in MLLMs, we also report the subset of attention contributions arising from image tokens. The DLA curves show that both logits rise sharply from around layer 20 onward and remain in competition until the final output. In layers 20--26, attention contributes strongly to the output logits, and image-token attention accounts for most of the attention-side DLA. Figure~\ref{fig:mechanistic_topk_head_attn_maps} further visualizes the attention maps of the top-4 and bottom-4 attention heads contributing to the bird--rabbit logit difference. These heads attend selectively to local image regions associated with one interpretation. This pattern is consistent with the observation of \citet{Kaduri2025-yw} that detailed visual information flows from image tokens to generated tokens mainly in the middle-to-late layers. In the final layers, MLP contributions become both positive and negative, suggesting a form of late-stage prediction ensembling \citep{Lad2025-gi}.

Motivated by prior findings that applying logit lens to image tokens at object locations can reveal the corresponding object token, and that ablating the embedding at that position makes the model unable to describe that object \citep{Neo2024-ep}, we next apply logit lens to image tokens. To better summarize the logit-lens results across layers, we use \emph{internal confidence} \citep{Jiang2024-mitigate}, defined for each token as the maximum probability attained across all layers, and classify a patch token as \emph{dominant} if this value exceeds a threshold. Following prior work, we use a threshold of 0.1 for \textit{LLaVA-1.5-7B}. Figure~\ref{fig:mechanistic_dominant_patch_map} shows an example dominant-patch map, in which the spatial distribution of dominant patches qualitatively overlaps with the image regions receiving high attention in Figure~\ref{fig:mechanistic_topk_head_attn_maps}. Because logit lens is only an observational method and does not by itself establish causal contribution to the output, we perform zero ablation on these embeddings. The probability of the ablated object's token in the output then drops to nearly zero, confirming that these tokens indeed contribute to reporting \texttt{"\_bird"} or \texttt{"\_rabb"}.

\subsection{How Does Modulability Occur?}

\paragraph{Hypothesis.}
The neutral analysis shows that selective attention to image tokens carrying duck- or rabbit-related evidence makes a large contribution to the final report. This suggests a simple working hypothesis: bottom-up modulation should primarily act by changing image-side evidence, especially the key/value content available to later attention, whereas top-down modulation may act by changing the text-side queries that retrieve interpretation-specific evidence from image tokens. In what follows, we test this hypothesis by examining how image-token representations vary under bottom-up manipulations and how cue information propagates through text tokens under top-down modulation.

\subsubsection{Bottom-up Modulation}

\begin{figure}
    \centering
    \includegraphics[width=\linewidth]{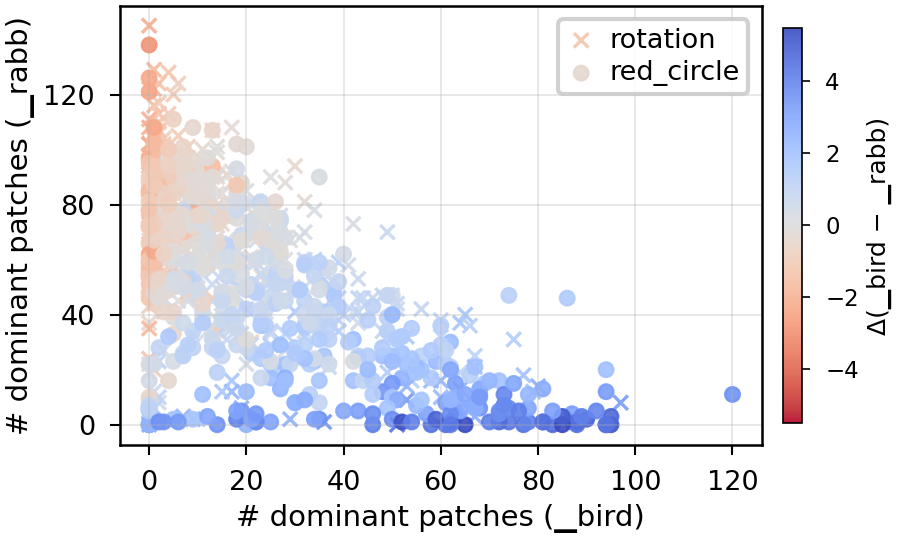}
    \caption{Bottom-up modulation tracks image-token evidence. Across rotation and red-circle conditions, the bird--rabbit logit difference broadly follows the balance of bird- and rabbit-dominant patches.}
    \label{fig:mechanistic_bottom_up_dpm_scatter}
\end{figure}

For bottom-up modulation, we examine how changes in the image are reflected in the image tokens and correlate with the final output. For both rotation---especially within the highly modulable range from $-5^\circ$ to $5^\circ$---and the red-circle manipulation, we plot the final logit difference against the difference in the number of dominant patch tokens (Figure~\ref{fig:mechanistic_bottom_up_dpm_scatter}). Comparing the number of bird-dominant and rabbit-dominant patches identified by logit lens, we find a trend that broadly matches the final output pattern. This supports the view that bottom-up modulation primarily changes the balance of image-side evidence available to downstream language-model processing.

\subsubsection{Top-down Modulation}

\begin{figure*}
    \centering
    \includegraphics[width=\textwidth]{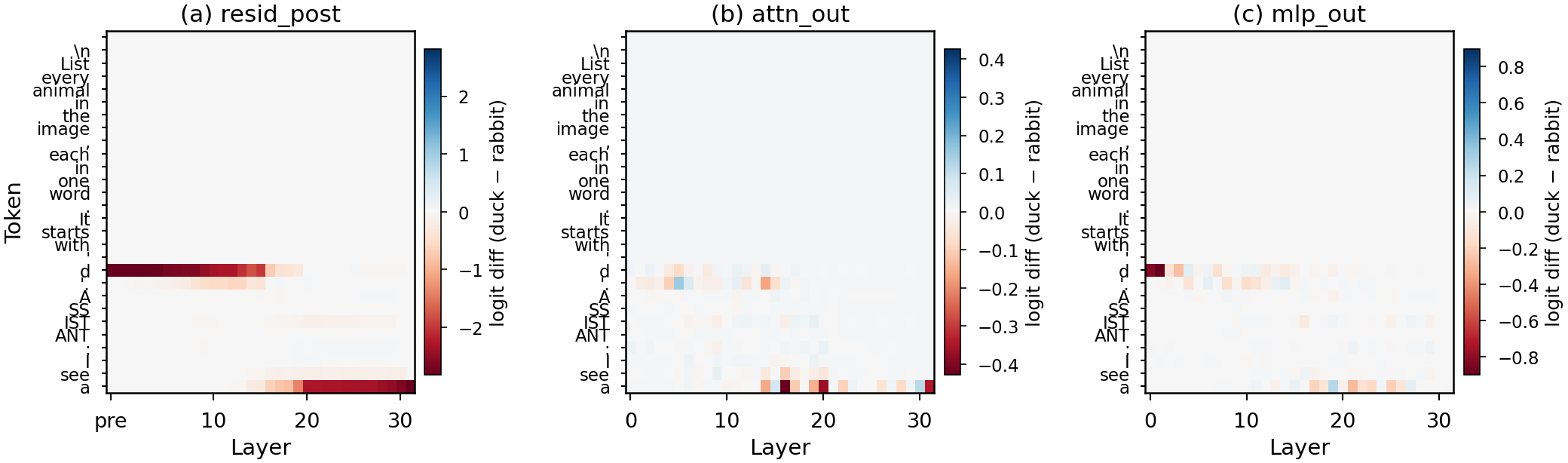}
    {\phantomsubcaption\label{fig:mechanistic_top_down_patching_every_resid_post}}
    {\phantomsubcaption\label{fig:mechanistic_top_down_patching_every_attn_out}}
    {\phantomsubcaption\label{fig:mechanistic_top_down_patching_every_mlp_out}}
    \caption{Layerwise resampling ablation for top-down cue runs in \textit{LLaVA-1.5-7B}, averaged over the model's valid Visual Anagram stimuli. We patch clean runs (e.g., \texttt{"It starts with 'd'."}) with corrupted runs (e.g., \texttt{"It starts with 'x'."}) and measure the effect on the final duck--rabbit logit difference.}
    \label{fig:mechanistic_top_down_patching_every}
\end{figure*}

\begin{figure}[t]
    \centering
    \includegraphics[width=\linewidth]{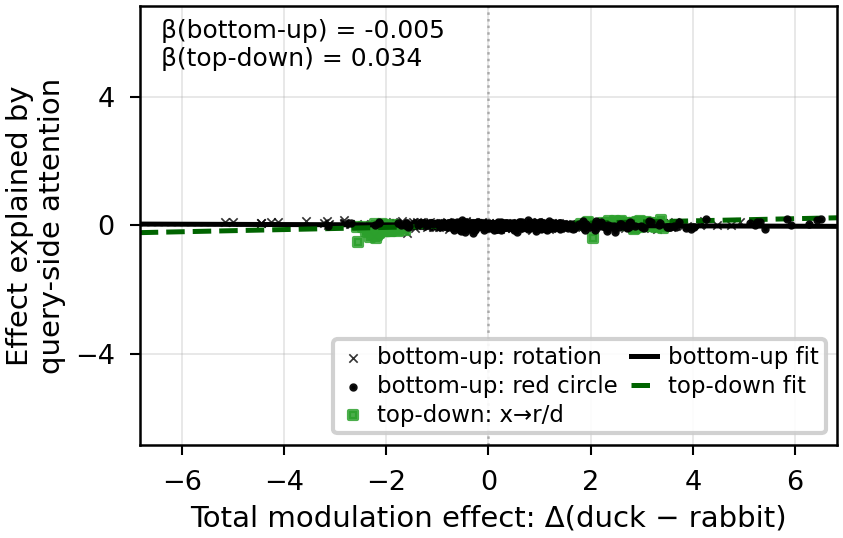}
    \caption{Query-side patching explains very little of the modulation effect. This suggests that neither bottom-up nor top-down modulation is primarily driven by changes in text-to-image attention queries.}
    \label{fig:mechanistic_image_attention_effect}
\end{figure}

We perform resampling ablation from corrupted top-down-modulated runs (e.g., \texttt{"It starts with 'x'."}) to clean top-down-modulated runs (e.g., \texttt{"It starts with 'r'/'d'."}) for each of \texttt{resid\_post}, \texttt{attn\_out}, and \texttt{mlp\_out} (Figure~\ref{fig:mechanistic_top_down_patching_every}). We find that the crucial cue information moves from the prefix token to the final token mainly over layers 15--20. Early-layer MLPs make especially strong indirect contributions to the final logits, resembling prior observations that early MLPs transform token-level information into coherent entities \citep{Elhage2022-ln,Gurnee2023-pk,Lad2025-gi,Kaplan2024-wl}. This was contrary to our initial query-mediated hypothesis: beyond this prefix-to-final-token transfer, the induced change in image-attention contribution is relatively small.

We conduct a more direct test of the earlier hypothesis by patching the query vectors of text tokens when computing similarities to image-token keys. We again patch clean top-down-modulated runs with corrupted runs and quantify how much of the total top-down effect is recovered. Across top-down-modulated runs (green points in Figure~\ref{fig:mechanistic_image_attention_effect}), this query-side intervention explains only about 3\% of the total effect.

\paragraph{Does bottom-up modulation lead to additional top-down modulation?}
Unlike top-down modulation, bottom-up modulation changes the image input directly while leaving the text unchanged. It could nevertheless induce an additional indirect effect through downstream text-token queries. To test this, we apply the same query-patching analysis to pairs of base runs and bottom-up-modulated runs with identical text. The black markers in Figure~\ref{fig:mechanistic_image_attention_effect} show that this effect is negligible, even relative to the already small effect observed in the top-down case.

Overall, the top-down effect appears to be introduced primarily through text-side processing rather than through substantial changes in crossmodal query-to-image matching.

\subsection{How Does Exclusivity Occur?}

\begin{figure}[t]
    \centering
    \includegraphics[width=\linewidth]{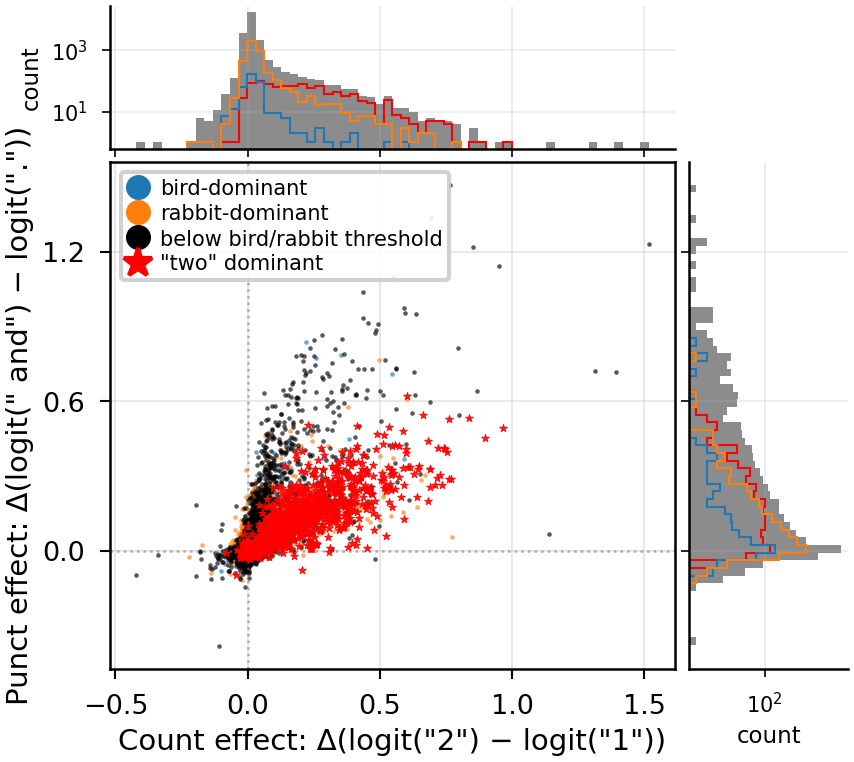}
    \caption{A small subset of image tokens affects both object-count encoding and exclusive reporting. Each point shows the effect of patching a single image-token embedding from the matched two-animal control into the bistable-image run.}
    \label{fig:mechanistic_exclusivity_count_vs_punct}
\end{figure}

\paragraph{Hypothesis.}
The behavioral analyses suggest a relationship between exclusivity and object count: the duck--rabbit is strongly reported as both exclusive and consisting of a single object, whereas both tendencies weaken for the Raven--Bear figure--ground stimulus. This motivates the hypothesis that object-count encoding contributes to exclusive reporting. We use causal interventions to test whether the computations supporting explicit object-count judgments overlap with those supporting exclusive continuation.

\paragraph{Analysis.}
We perform single-token image-embedding patching from the matched two-animal control run into the bistable-image run. For each patched token, we measure its effect on the object-count contrast, \texttt{"\_2"} $-$ \texttt{"\_1"}, and on an exclusivity-related continuation contrast, \texttt{"\_and"} $-$ \texttt{"."}. As shown in Figure~\ref{fig:mechanistic_exclusivity_count_vs_punct}, a small subset of image tokens has large positive effects on both measures, supporting a shared causal substrate linking object-count processing and exclusive reporting. Interestingly, these tokens are largely distinct from the dominant patches of \texttt{"\_bird"} or \texttt{"\_rabb"}, and most \texttt{"\_2"}-dominant patches have little individual effect on either measure, although some do; this is consistent with the observation that number-like tokens read out from image positions by the logit lens often reflect language-model processing rather than causally relevant visual representations \citep{Neo2024-ep,Neo2026-fw}.
\section{Conclusion}
We studied whether MLLMs exhibit human-like report behavior on bistable images, focusing on \emph{modulability} and \emph{exclusivity}. Across the duck--rabbit and synthetic Visual Anagram stimuli, both bottom-up visual manipulations and top-down linguistic cues systematically biased reported interpretations, showing modulability, while responses remained predominantly exclusive.

Mechanistically, these effects were linked to competing image-token representations, distinct pathways for bottom-up and top-down modulation, and a close relation between exclusive reporting and object-count encoding. These findings show that human-like bistable-image reporting can emerge in MLLMs in computationally interpretable forms, and they support the use of bistable images as a compact testbed for studying how ambiguous visual evidence is transformed into language.

\section*{Limitations}

\paragraph{Scope of modulability and exclusivity.}
Our analysis focuses on two report-level properties of bistable-image processing: \emph{modulability}, whether reported interpretations can be biased by bottom-up and top-down factors, and \emph{exclusivity}, whether the model commits to a single interpretation in a given response. These properties are not specific to bistable images: modulability reflects the influence of sensory evidence and contextual factors on perceptual interpretation, while exclusivity reflects the tendency to integrate conflicting local evidence into a coherent object-level interpretation, consistent with hierarchical accounts of perception \citep{Weilnhammer2017-xe,Parr2019-bp}. Bistable images make these properties especially tractable, but they do not exhaust human bistable perception or perception more generally.

\paragraph{Switching dynamics and active visual sampling.}
A central limitation is that our experiments do not address the temporal switching dynamics characteristic of human bistable perception. In humans, a fixed ambiguous stimulus can alternate between interpretations over continued viewing, shaped by factors including eye movements, adaptation, and selective fixation \citep{Brascamp2018-gq,Hsu2025-sj}. Here, by contrast, each MLLM encodes a fixed image and produces a linguistic report. Report changes elicited by re-prompting or stochastic decoding would therefore not by themselves constitute human-like perceptual switching. A closer model-side analogue would require active-perception mechanisms that acquire new visual evidence over time through actions such as shifting fixation, cropping, or zooming. Active-perception MLLMs such as DeepEyes \citep{Zheng2025-hx} provide a concrete future testbed: unlike our externally imposed red-circle cue, such models can select new visual evidence themselves, allowing switching-like dynamics and their mechanisms to be studied in a general-purpose learned system.

\paragraph{Generalization beyond object-category bistability.}
Our primary experiments concern static object-category ambiguity, where one image region supports incompatible object identities. Other forms of bistable or multistable perception, including depth, figure--ground, motion, and binocular rivalry, differ in the structure of their competing interpretations and may require different operationalizations \citep{Leopold1999-ae,Brascamp2018-gq,Meng2004-zh}. Our Raven--Bear analysis provides an initial figure--ground boundary condition: modulability generalizes, but exclusivity is weaker and covaries with whether the model construes the image as containing one or two objects (Section~\ref{sec:behavioral_exclusivity}; Appendix~\ref{sec:appendix_figure_ground}). Systematic generalization across other forms of bistability remains for future work. We discuss further extensions, including history dependence and relational ambiguity, in Appendix~\ref{sec:additional_discussion}.

\paragraph{Interpretive scope.}
More broadly, our comparisons concern report behavior and the computational mechanisms that support it. Although these dimensions provide useful points of comparison between humans and MLLMs, similarities along them should not be taken as evidence for corresponding similarities in perceptual experience or consciousness.

\section*{Acknowledgments}
We thank Taiga Shinozaki for helpful discussions. This work was supported by JSPS KAKENHI Grant Number JP24H00809, and JST CREST Grant Number JPMJCR2565, Japan. We used generative AI tools to assist with manuscript preparation and code development. All AI-assisted outputs were reviewed and verified by the authors, who take full responsibility for all aspects of this work.

\bibliography{main}

\appendix
\label{sec:appendix}

\section{Details of General Setup}
\label{sec:appendix_setup}

\subsection{Stimuli and preprocessing}

\subsubsection{Visual Anagrams and matched controls}
\label{sec:appendix_visual_anagram}

\paragraph{Candidate generation.}
Following the original Visual Anagram framework \citep{Geng2024-be}, we sampled candidates from DeepFloyd IF (\texttt{IF-I-M-v1.0}, \texttt{IF-II-M-v1.0}, followed by the Stable Diffusion $\times 4$ upscaler; 30 inference steps, guidance scale 10.0) with the views \texttt{['identity', 'rotate\_cw']} and the prompts \texttt{"drawing of a duck head"} (identity view) and \texttt{"drawing of a rabbit head"} (rotated view). We generated 100 candidates from different random seeds, keeping the $1024\times1024$ identity view. Each candidate's background was removed automatically (rembg with u2net), and the result was pre-rotated by $-45^\circ$ and cropped to its alpha bounding box, so that the duck$\leftrightarrow$rabbit flip spans approximately $\pm45^\circ$ in the rotation experiments.

\begin{figure*}[t]
    \centering
    \includegraphics[width=\textwidth]{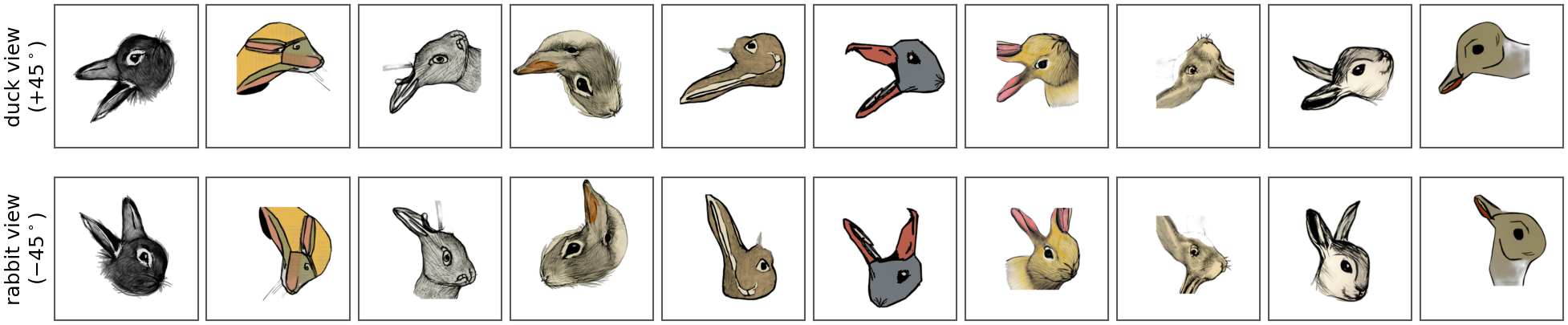}
    \caption{Ten randomly selected stimuli from the 50-stimulus synthetic Visual Anagram set, rendered as in the experiments. Top: duck view ($+45^\circ$); bottom: rabbit view ($-45^\circ$).}
    \label{fig:appendix_va_gallery}
\end{figure*}

\paragraph{Judge-based validity filter.}
Stimulus validity was assessed by an external VLM judge (\texttt{gpt-5-nano}), independently of the models under study. Each candidate was rendered exactly as in the experiment pipeline (composited on white on a canvas padded to the image diagonal and resized to $336\times336$) at the duck view ($+45^\circ$) and rabbit view ($-45^\circ$). Each view was queried five times in independent contexts with \texttt{"What animal is in this drawing? Answer with a single word."}; answers were mapped to a duck class (\textit{duck, duckling, mallard, goose, bird, waterfowl, swan, pelican}) or a rabbit class (\textit{rabbit, bunny, hare}). A candidate was accepted if at least four of five answers were duck-class at the duck view and at least four of five were rabbit-class at the rabbit view. Fifty of the 100 candidates passed this criterion and constitute the final stimulus set; Figure~\ref{fig:appendix_va_gallery} shows examples.

\paragraph{Matched non-ambiguous controls.}
For each accepted stimulus, we generated a style-matched \emph{two-animal control} showing a duck and a rabbit side by side on a plain background (\texttt{gpt-image-1-mini} image editing, conditioned on the stimulus rendered in experiment geometry). From each two-animal control, we derived duck-only and rabbit-only \emph{single-animal controls} by zeroing the alpha channel of the other animal on the same canvas. All 100 resulting single-animal controls passed the same judge check.

All images used in our analyses are resized to $336\times336$ before being passed to the model processor.

\subsubsection{Red-circle manipulation}
\label{sec:appendix_red_circle}

\begin{figure}[t]
    \centering
    {\setlength{\fboxsep}{0pt}\setlength{\fboxrule}{0.4pt}\fbox{\includegraphics[width=0.8\linewidth]{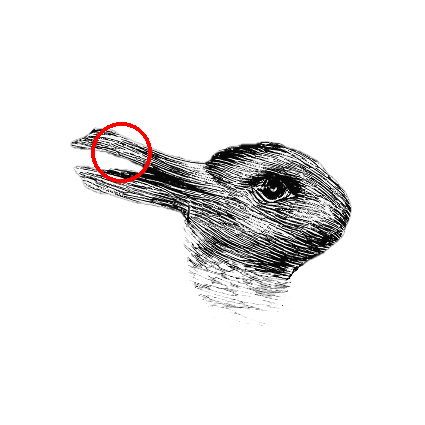}}}
    \caption{Illustration of the red-circle manipulation. A fixed-radius red circle is swept across Harper's version on a regular grid, and the model output is recorded at each circle center.}
    \label{fig:harper-rc-left}
\end{figure}

For the red-circle manipulation, we use Harper's version of the duck--rabbit resized to $336\times336$. A red circle with line width 4 pixels and radius 24 pixels is swept across the image on a regular grid with stride 48 pixels in both directions. At each position, we rerun the model and record the logits of the relevant interpretation tokens. Figure~\ref{fig:harper-rc-left} illustrates the manipulation.

\subsection{Evaluation details}

\subsubsection{Decision-boundary fitting and per-model inclusion}
\label{sec:appendix_boundary_fit}

Analyses that center each Visual Anagram on a model-specific decision boundary require a well-constrained boundary estimate. For each model--stimulus pair, we fit a logistic curve to the duck-vs.-rabbit first-token probabilities across rotation angles and define the decision boundary as its 50\% point. A fit is rejected if fewer than five angles are usable, the predicted probability range within the sweep is below $0.15$, the maximal slope is below $0.002$ per degree, or fewer than three points lie on each side of the $0.4/0.6$ thresholds. Model--stimulus pairs with rejected fits are excluded from analyses that align or evaluate stimuli relative to the decision boundary. Table~\ref{tab:appendix_va_valid_n} reports the number of included stimuli per model under the default prompt.

\begin{table}[t]
\centering
\small
\begin{tabular}{lc}
\toprule
Model & Valid stimuli (of 50) \\
\midrule
LLaVA-1.5-7B & 44 \\
LLaVA-1.5-13B & 42 \\
LLaVA-v1.6-Vicuna-7B & 40 \\
LLaVA-v1.6-Mistral-7B & 40 \\
Llama3-LLaVA-Next-8B & 41 \\
Qwen2-VL-7B-Instruct & 41 \\
SmolVLM-Instruct & 40 \\
IDEFICS2-8B & 40 \\
InstructBLIP-Vicuna-7B & 37 \\
\bottomrule
\end{tabular}
\caption{Number of Visual Anagram stimuli with a valid decision-boundary fit per model (default prompt).}
\label{tab:appendix_va_valid_n}
\end{table}

\subsubsection{Classifying reports by the animals they name}
\label{sec:appendix_beam_classification}

For the aggregate exclusivity analyses, we run beam search with 8 beams, up to 12 new tokens, and no length penalty. Each returned continuation is classified with an LLM judge (GPT-5-nano). The judge lists the animals named in the continuation, including the prefix, and each listed noun is mapped to bird-like, rabbit-like, or other. A continuation is then labeled \emph{single} if it names one animal, \emph{multiple} if it names two or more distinct animals, or \emph{none} if it names no animal. We report each class's share of the probability mass of the returned beams, averaged across stimuli (mean $\pm$ SD). Continuation probabilities exclude the end-of-sequence term, which is an unnatural stop for some tokenizers.

The same scheme classifies the free-form generations of Appendix~\ref{sec:appendix_across_prompts} through a lexicon distilled from the judge: the animal nouns it extracted from 133k distinct generations, each mapped once to bird-like, rabbit-like, or other, keeping nouns listed at least 20 times and discarding those naming no species (\texttt{animal}, \texttt{creature}) or a body part (\texttt{beak}, \texttt{ears}). The resulting 103 nouns are matched on word boundaries. All bird-like nouns are treated as referring to the same animal, as are all rabbit-like nouns; for example, ``a small bunny \ldots\ the rabbit'' is classified as naming one animal, whereas ``a rabbit and a cat'' is classified as naming two. On the judged generations this lexicon reproduces the judge's own labels for 97.4\% of 207k texts.

\section{Additional Behavioral Results}
\label{sec:appendix_behavioral}

\subsection{Across LLaVA-family models}
\label{sec:appendix_llava_all_models}
\subsubsection{Rotation}
\label{sec:appendix_rotation_all_models}

\begin{figure*}[t]
    \centering
    \includegraphics[width=\textwidth]{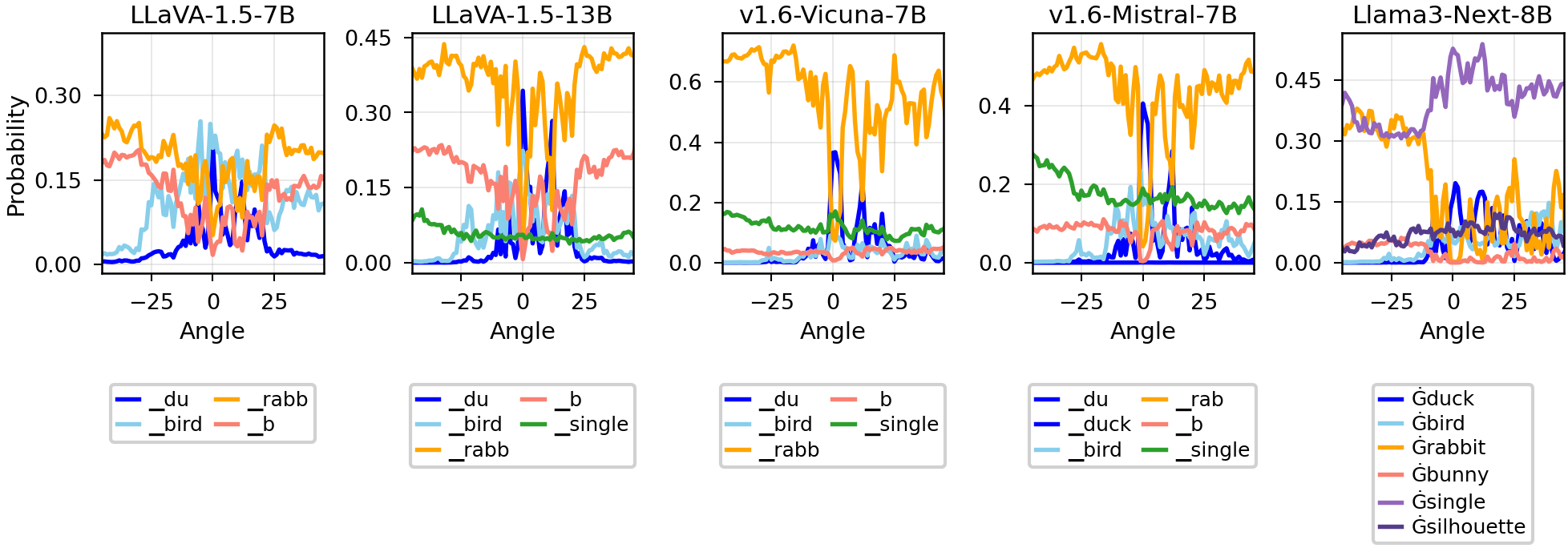}
    \caption{Rotation-based modulability across all models for Harper's version. All models show a duck-dominant interpretation near $0^\circ$, but the sharpness and location of the reversal differ by model.}
    \label{fig:appendix_rotation_harper_all_models}
\end{figure*}

\begin{figure*}[t]
    \centering
    \includegraphics[width=\textwidth]{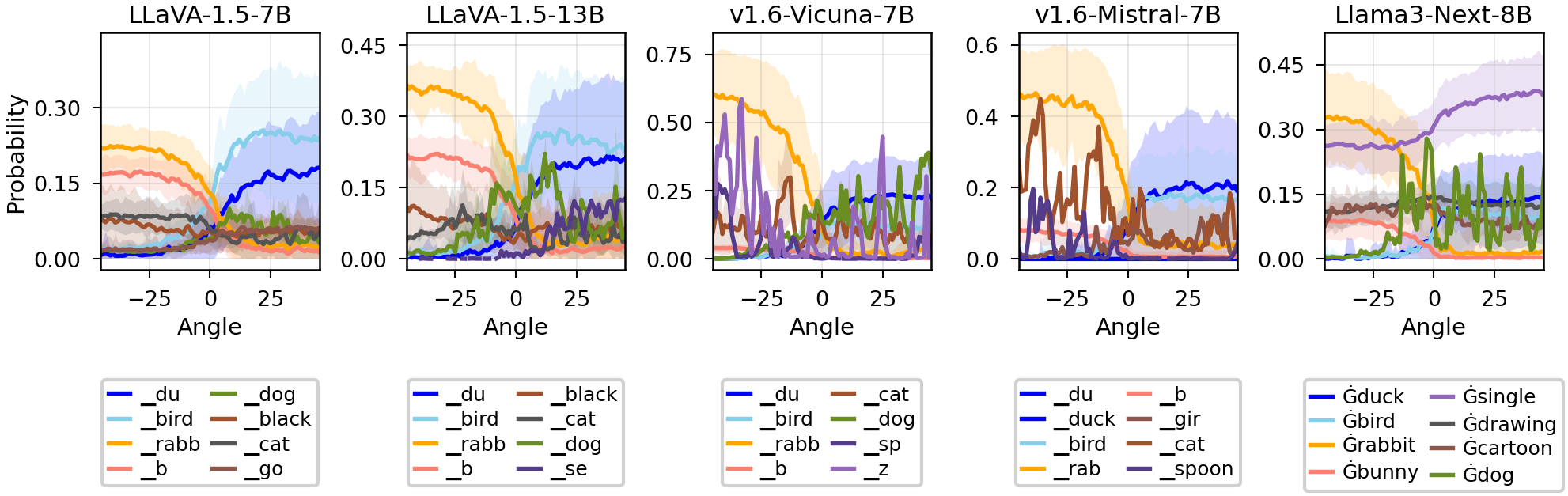}
    \caption{Rotation-based modulability across all models for the aggregated Visual Anagram stimuli. The same qualitative reversal pattern appears across models, though the reversal profile varies in sharpness and stability.}
    \label{fig:appendix_rotation_va_all_models}
\end{figure*}

Figures~\ref{fig:appendix_rotation_harper_all_models} and \ref{fig:appendix_rotation_va_all_models} show rotation-based modulability across all five LLaVA-family models. For Harper's version, duck-related tokens are preferred over rabbit-related tokens near $0^\circ$ in every model, although the sharpness and location of the reversal vary substantially. The aggregated Visual Anagrams likewise show rotation-dependent shifts in the relative bird--rabbit preference across models, with model-specific differences in the reversal profile and in the probability assigned to competing output tokens.

\subsubsection{Red Circle}
\label{sec:appendix_red_circle_all_models}

\begin{figure*}[t]
    \centering
    \includegraphics[width=\textwidth]{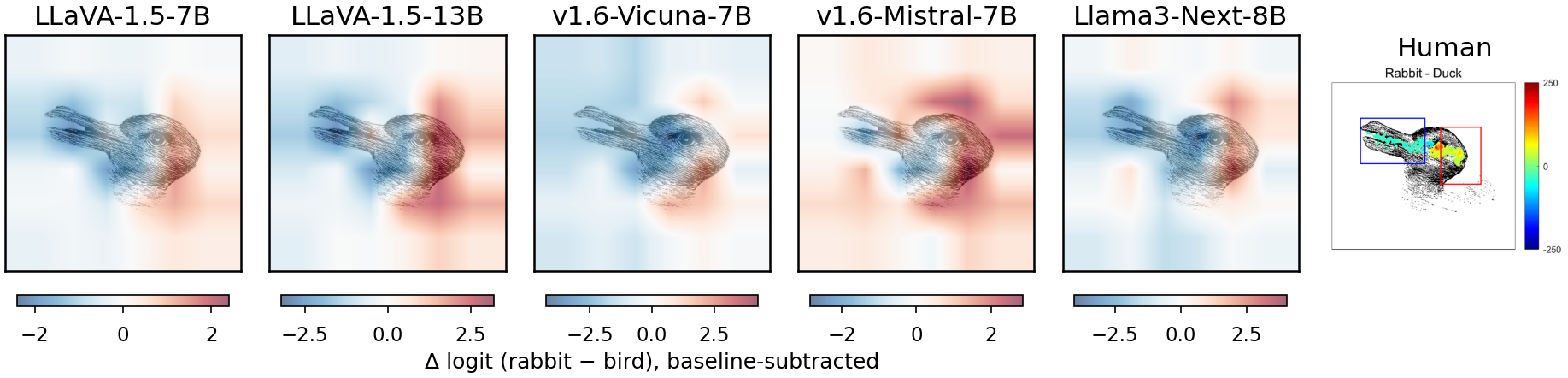}
    \caption{Red-circle modulation on Harper's version across LLaVA-family models: rabbit-vs-duck difference maps (as in Figure~\ref{fig:behavioral_red_circle}), with the human fixation-density map \citep{Hsu2025-sj} shown for reference (right).}
    \label{fig:appendix_red_circle_all_models}
\end{figure*}

Figure~\ref{fig:appendix_red_circle_all_models} shows the red-circle manipulation across all LLaVA-family models. Although the sharpness of the maps varies, all five models show a broadly similar spatial dissociation between duck-supporting and rabbit-supporting regions, with the strongest rabbit-related effects near the mouth-like region and the strongest duck-related effects near the beak-like region.

\subsubsection{Top-down cues}
\label{sec:appendix_topdown_all_models}

\begin{figure*}[t]
    \centering
    \includegraphics[width=\textwidth]{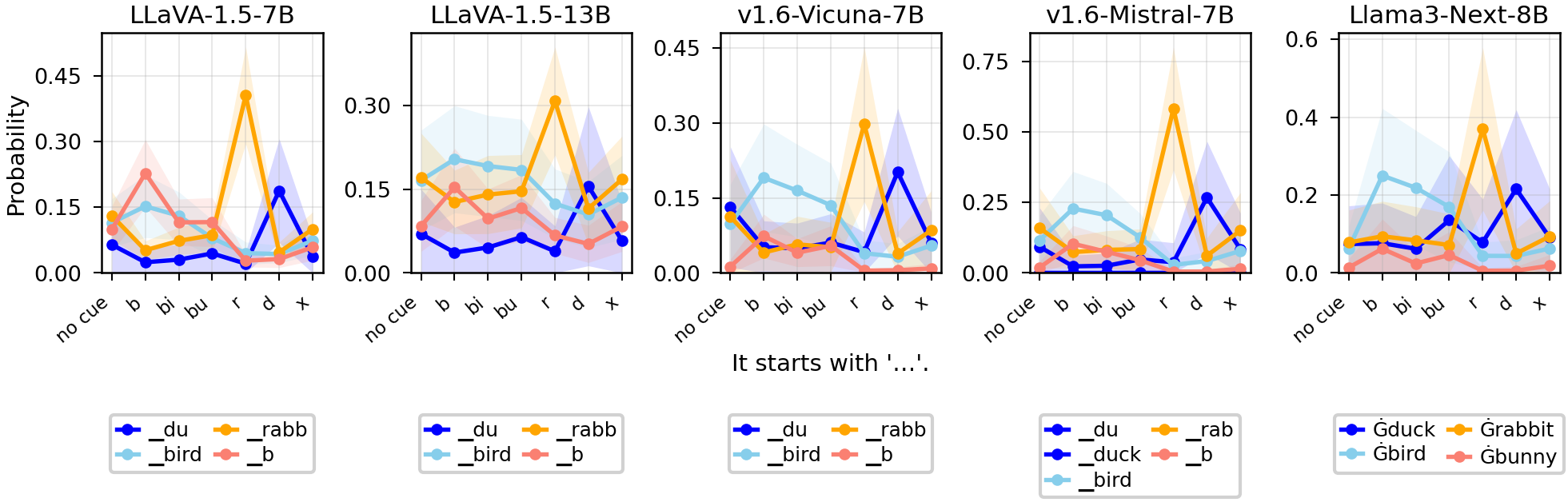}
    \caption{Top-down modulation across all models for prefix cues. Prefix cues consistently bias the output toward the intended interpretation.}
    \label{fig:appendix_topdown_prefix_all_models}
\end{figure*}

\begin{figure*}[t]
    \centering
    \includegraphics[width=\textwidth]{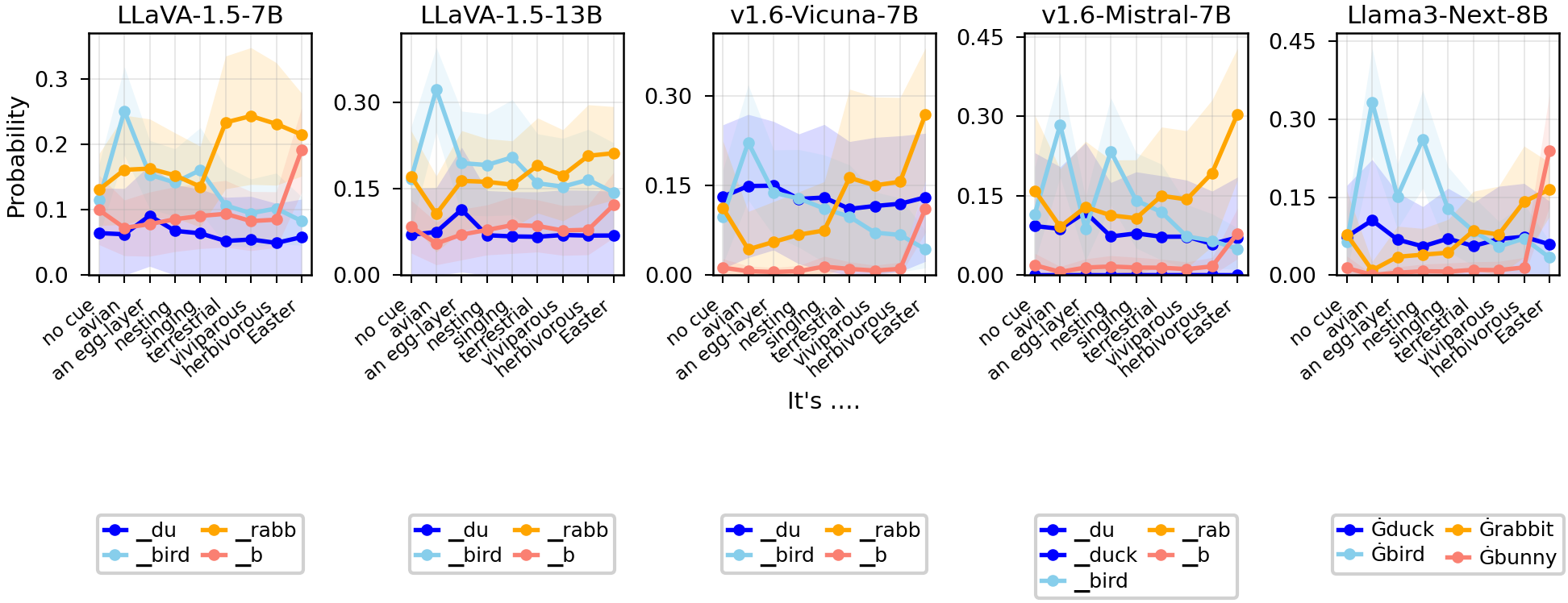}
    \caption{Top-down modulation across all models for semantic cues. The qualitative effect is similar to the prefix condition, but weaker.}
    \label{fig:appendix_topdown_semantic_all_models}
\end{figure*}

Figures~\ref{fig:appendix_topdown_prefix_all_models} and \ref{fig:appendix_topdown_semantic_all_models} show top-down cue effects across the five LLaVA-family models. Prefix cues produce strong and consistent shifts toward the intended interpretation, whereas semantic cues produce weaker and more model-dependent shifts.

\subsubsection{Exclusivity under beam search}
\label{sec:appendix_exclusivity_all_models}

\begin{figure*}[t]
    \centering
    \includegraphics[width=\textwidth]{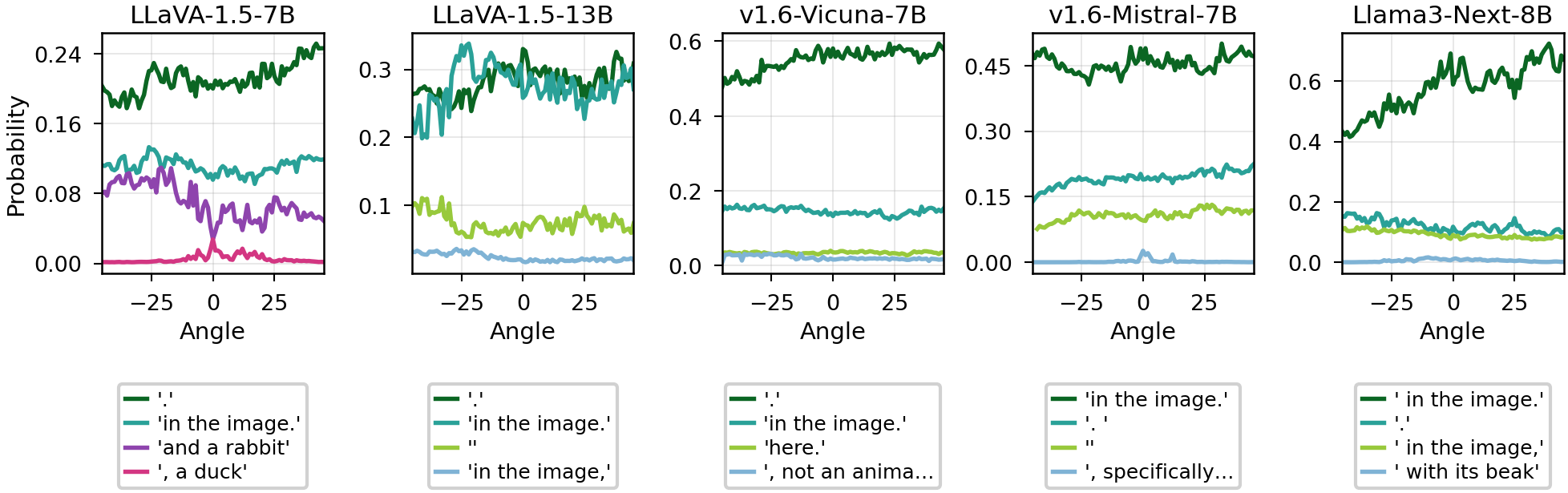}
    \caption{Beam-search results across all models for the bird-prefixed condition. In all models, single-interpretation continuations are preferred to continuations that would naturally enumerate both animals.}
    \label{fig:appendix_beam_bird_all_models}
\end{figure*}

\begin{figure*}[t]
    \centering
    \includegraphics[width=\textwidth]{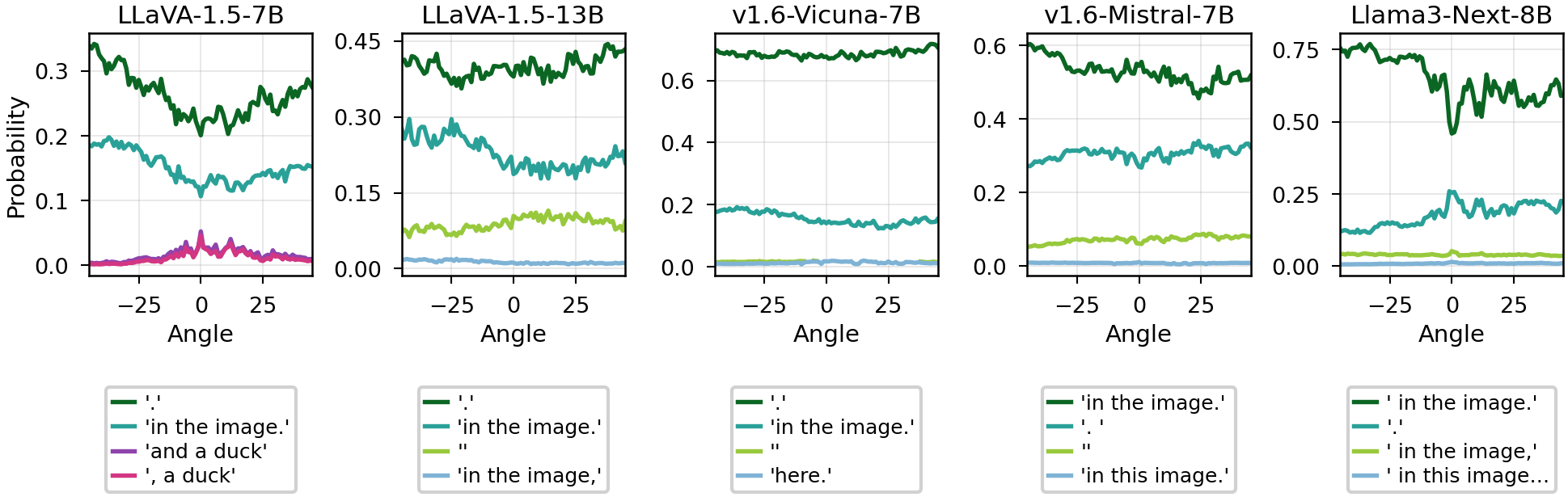}
    \caption{Beam-search results across all models for the rabbit-prefixed condition. The same qualitative exclusivity pattern is observed when the opposite interpretation is forced into the prefix.}
    \label{fig:appendix_beam_rabbit_all_models}
\end{figure*}

\begin{figure*}[t]
    \centering
    \includegraphics[width=\textwidth]{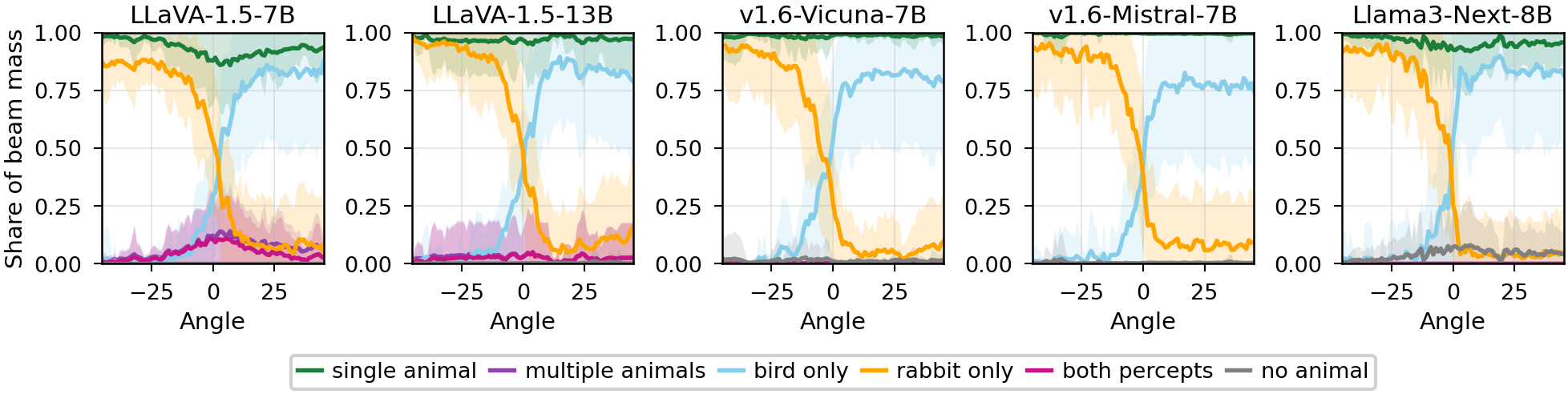}
    \caption{Beam-continuation classes on aggregated Visual Anagrams under the default prefix, across LLaVA-family models (mean $\pm$ SD across stimuli).}
    \label{fig:appendix_beam_va_all_models}
\end{figure*}

Figures~\ref{fig:appendix_beam_bird_all_models} and \ref{fig:appendix_beam_rabbit_all_models} show beam-search continuations on Harper's version when either interpretation is forced into the prefix. Across models, continuations that terminate or elaborate the prefixed interpretation are preferred over continuations that enumerate the competing interpretation. Figure~\ref{fig:appendix_beam_va_all_models} shows the corresponding aggregate over the Visual Anagrams under the default prefix: multiple-animal continuations remain rare across models, accounting for 0--12\% of beam mass at the ambiguity boundary (Table~\ref{tab:appendix_object_count}).

\subsection{Behavioral checks on non-LLaVA models}
\label{sec:appendix_non_llava}

To test whether the behavioral findings are specific to the LLaVA family, we run focused behavioral checks on four non-LLaVA models: \textit{Qwen2-VL-7B-Instruct}, \textit{SmolVLM-Instruct}, \textit{IDEFICS2-8B}, and \textit{InstructBLIP-Vicuna-7B}. We use the same operationalization as in the main experiments, testing rotation-based bottom-up modulation, red-circle modulation, top-down linguistic modulation, and exclusivity under beam search. Overall, the qualitative pattern of modulability and predominantly exclusive reporting extends beyond the LLaVA family, although effect sizes and reversal profiles vary across architectures.

\subsubsection{Rotation}
\label{sec:appendix_non_llava_rotation}

\begin{figure*}[t]
    \centering
    \includegraphics[width=\textwidth]{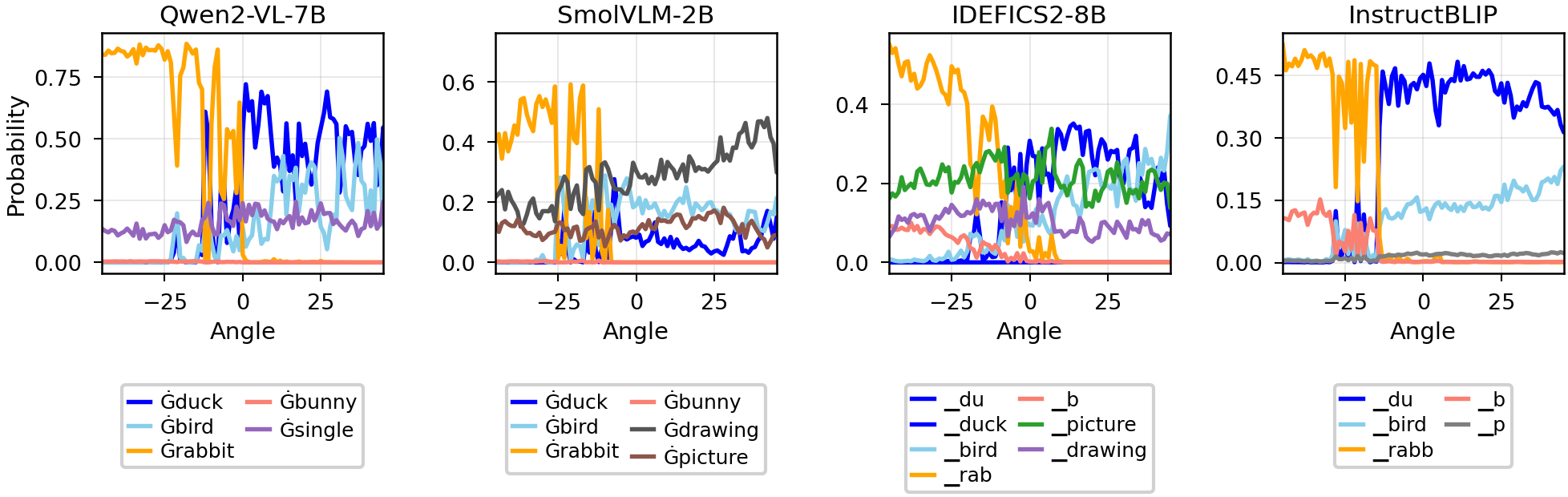}
    \caption{Rotation-based modulation on Harper's duck--rabbit across non-LLaVA models.}
    \label{fig:appendix_non_llava_rotation_harper}
\end{figure*}

\begin{figure*}[t]
    \centering
    \includegraphics[width=\textwidth]{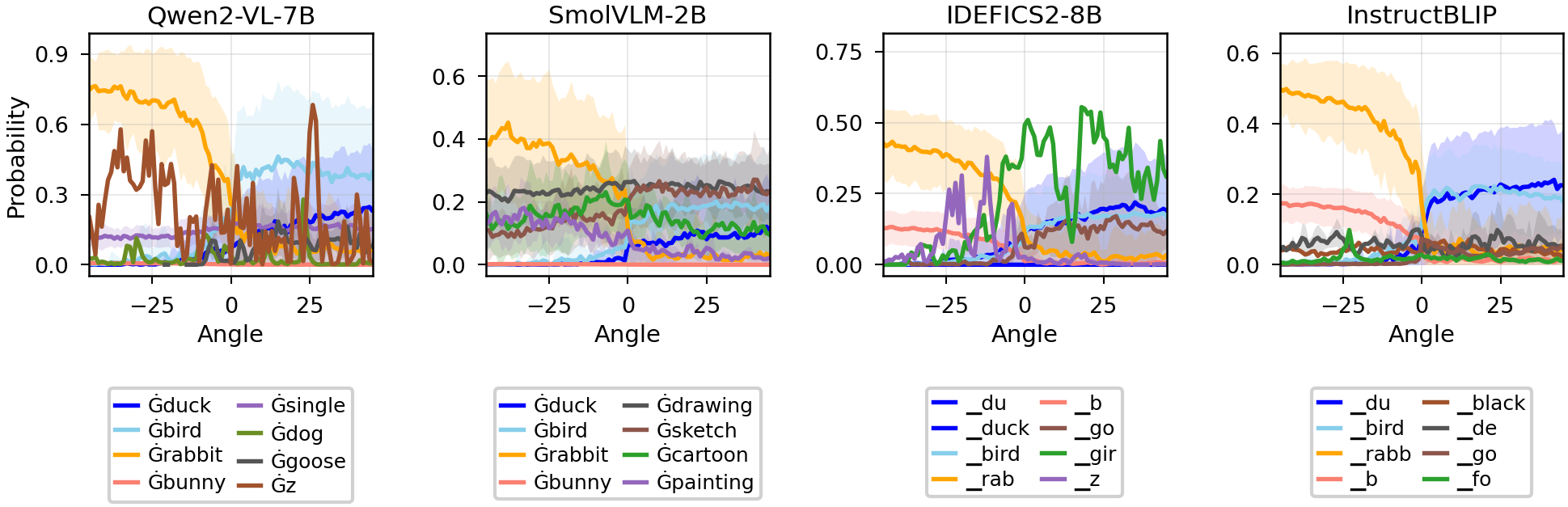}
    \caption{Rotation-based modulation on aggregated Visual Anagram stimuli across non-LLaVA models.}
    \label{fig:appendix_non_llava_rotation_va}
\end{figure*}

Figures~\ref{fig:appendix_non_llava_rotation_harper} and \ref{fig:appendix_non_llava_rotation_va} show rotation-based modulation across the four non-LLaVA models. All models show rotation-dependent shifts in the relative duck/bird- and rabbit-related outputs on both Harper's version and the Visual Anagrams, although the profiles are more heterogeneous than within the LLaVA family and some models assign substantial probability to competing non-target tokens.

\subsubsection{Red-circle modulation}
\label{sec:appendix_non_llava_red_circle}

\begin{figure*}[t]
    \centering
    \includegraphics[width=\textwidth]{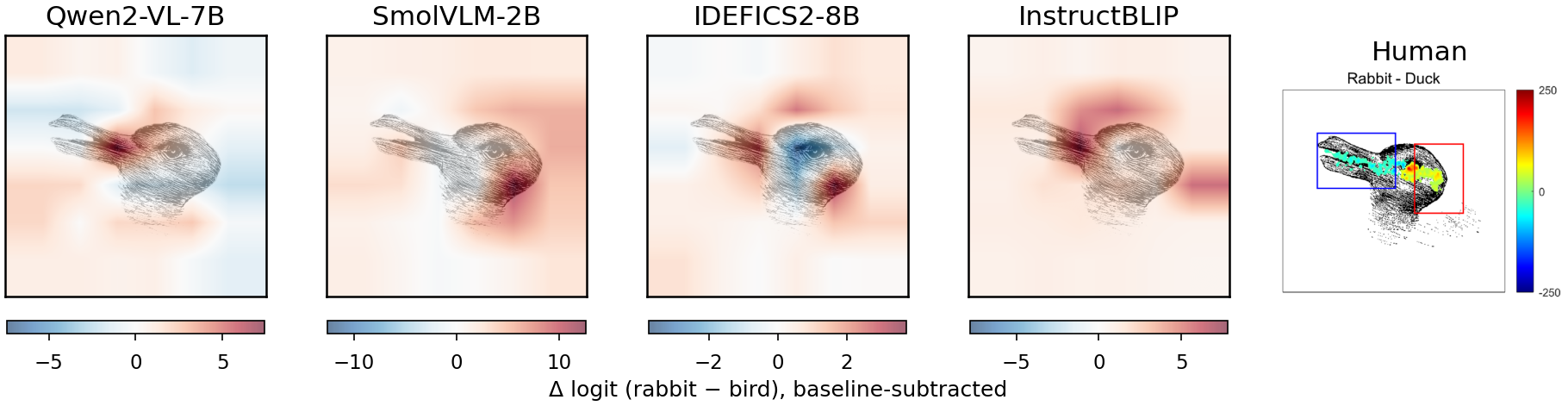}
    \caption{Red-circle modulation on Harper's version across non-LLaVA models: rabbit-vs-duck difference maps, with the human fixation-density map \citep{Hsu2025-sj} shown for reference (right).}
    \label{fig:appendix_non_llava_red_circle}
\end{figure*}

Figure~\ref{fig:appendix_non_llava_red_circle} shows red-circle modulation across the non-LLaVA models. In contrast to the consistent spatial dissociation observed within the LLaVA family, the non-LLaVA models vary in spatial selectivity: cues near the rabbit's mouth still shift \textit{SmolVLM} and \textit{IDEFICS2} toward rabbit-related outputs, while the maps of \textit{Qwen2-VL-7B} and \textit{InstructBLIP} align less clearly with that arrangement. The red-circle cue thus modulates the outputs of all models, but does not reproduce the same spatial dissociation in every architecture. The relatively consistent red-circle effect across LLaVA variants may partly reflect their shared CLIP ViT-L/14@336 vision encoder, for which red-circle overlays have been shown to provide effective zero-shot visual prompting; standard LLaVA models also exhibit visual-prompt sensitivity without prompt-specific training \citep{Shtedritski2023-ai,Cai2024-on}. Comparable evidence for this specific cue is lacking for the non-LLaVA architectures evaluated here, and visual-prompt effectiveness is known to vary across models and prompt configurations \citep{Xu2026-nv}, which may contribute to their more heterogeneous effects.

\subsubsection{Top-down cues}
\label{sec:appendix_non_llava_topdown}

\begin{figure*}[t]
    \centering
    \includegraphics[width=\textwidth]{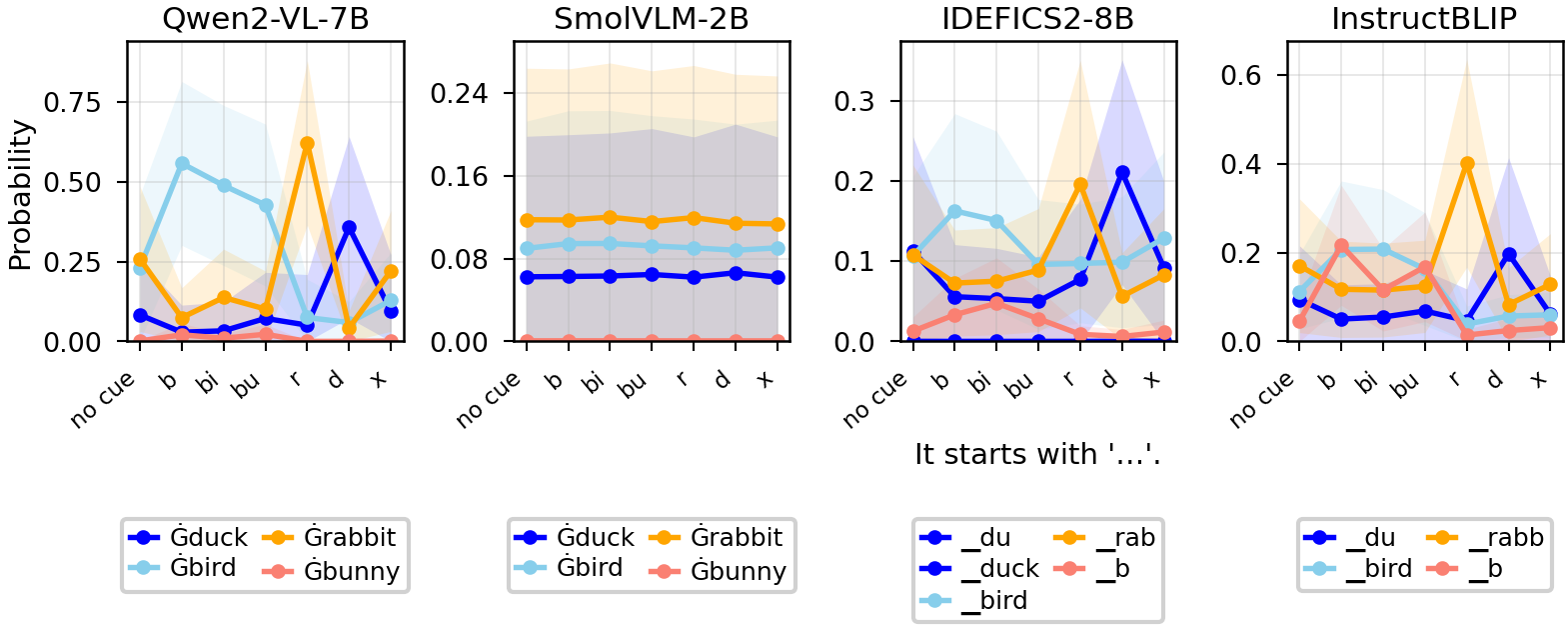}
    \caption{Top-down modulation by prefix cues across non-LLaVA models.}
    \label{fig:appendix_non_llava_topdown_prefix}
\end{figure*}

\begin{figure*}[t]
    \centering
    \includegraphics[width=\textwidth]{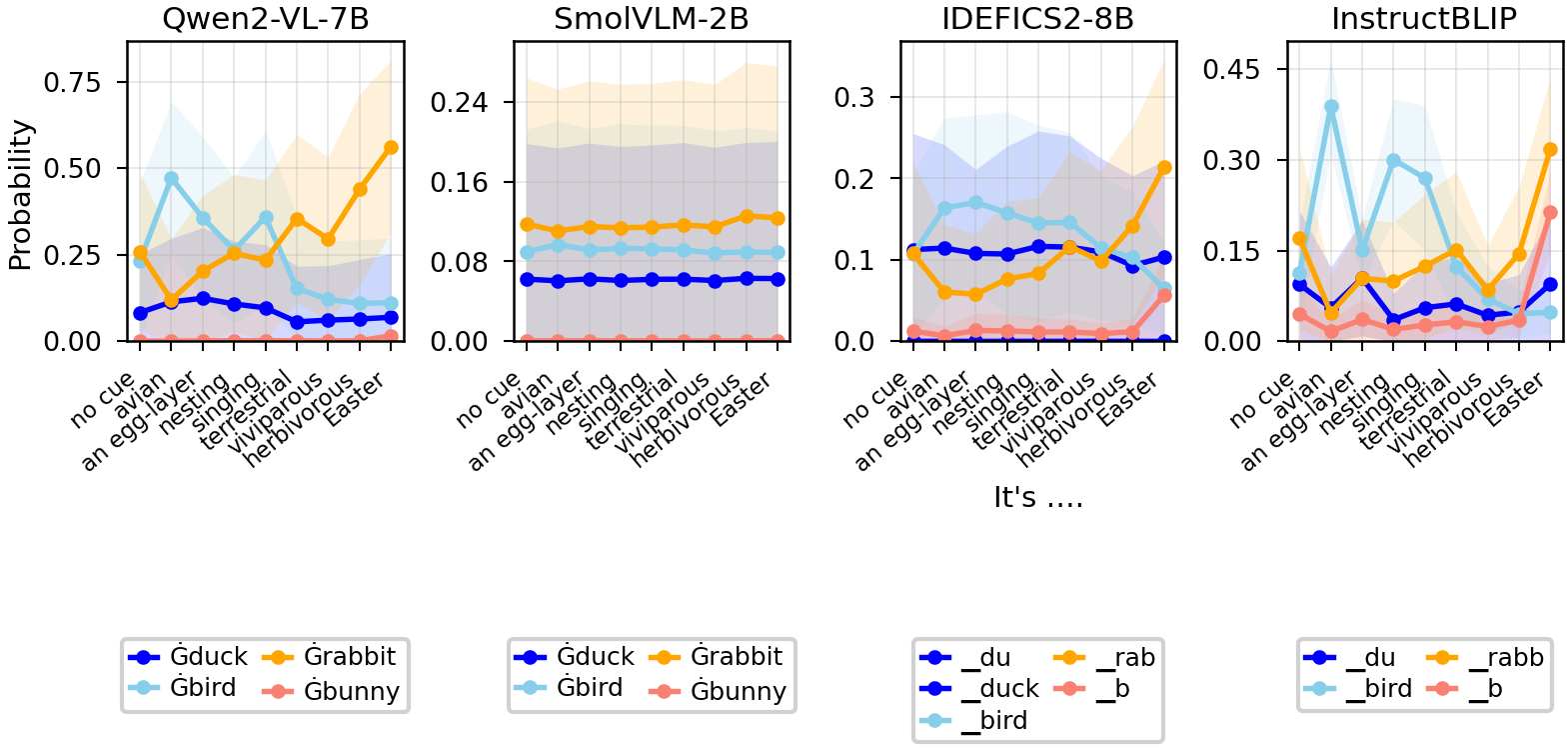}
    \caption{Top-down modulation by semantic cues across non-LLaVA models.}
    \label{fig:appendix_non_llava_topdown_semantic}
\end{figure*}

Figures~\ref{fig:appendix_non_llava_topdown_prefix} and \ref{fig:appendix_non_llava_topdown_semantic} show top-down cue effects across the non-LLaVA models. Prefix cues generally produce stronger effects than semantic cues, but susceptibility varies substantially by model. \textit{SmolVLM} is the clearest outlier: even its strongest prefix cue changes the rabbit-related probability by less than 0.01, compared with roughly 0.1--0.4 in the other models.

This weak top-down susceptibility may relate to the unusually weak task-dependent visual integration independently reported for SmolVLM by \citet{Long2025-we}. Their analysis considers visual-to-language influence rather than the language-to-visual modulation studied here, but both results point to relatively weak context-dependent cross-modal interaction in this model.

\subsubsection{Exclusivity under beam search}
\label{sec:appendix_non_llava_beam}

\begin{figure*}[t]
    \centering
    \includegraphics[width=\textwidth]{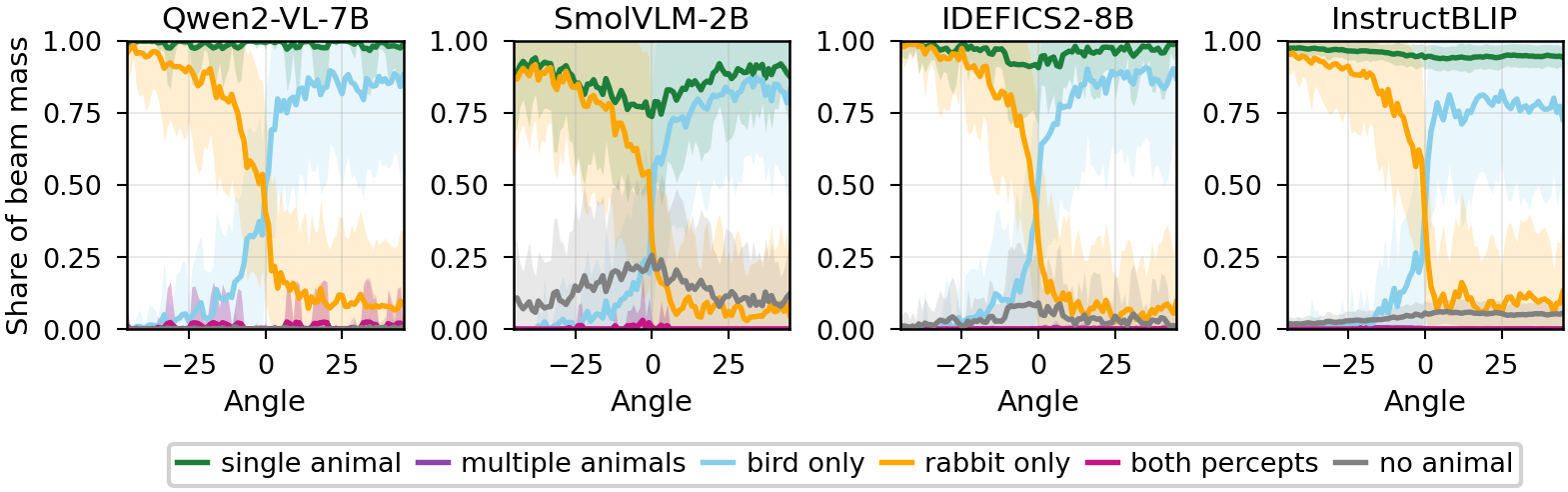}
    \caption{Beam-continuation classes on aggregated Visual Anagrams under the default prefix, across non-LLaVA models (mean $\pm$ SD across stimuli).}
    \label{fig:appendix_non_llava_beam_va}
\end{figure*}

Figure~\ref{fig:appendix_non_llava_beam_va} shows beam-continuation classes on the aggregated Visual Anagrams under the default prefix. As in the LLaVA family, multiple-animal continuations remain near zero across all four models, and single-animal continuations hold most of the beam mass. \textit{SmolVLM} additionally assigns a noticeable fraction of beam mass to continuations that name no animal, but does not preferentially enumerate both interpretations.

\subsection{Matched non-ambiguous controls}
\label{sec:appendix_control_images}

We use the matched controls defined in Appendix~\ref{sec:appendix_visual_anagram} to test whether the behavioral effects are specific to ambiguous stimuli. Two-animal controls, containing a duck and rabbit as separate objects, test rotation and exclusivity; single-animal controls test whether top-down cues can induce an interpretation unsupported by the image. Each control is paired with a Visual Anagram stimulus, and analyses include only pairs for which that stimulus has a valid model-specific decision boundary.

\subsubsection{Rotation and exclusivity on two-animal controls}
\label{sec:appendix_control_rotation}
\label{sec:appendix_control_beam}

\begin{figure*}[t]
    \centering
    \includegraphics[width=\textwidth]{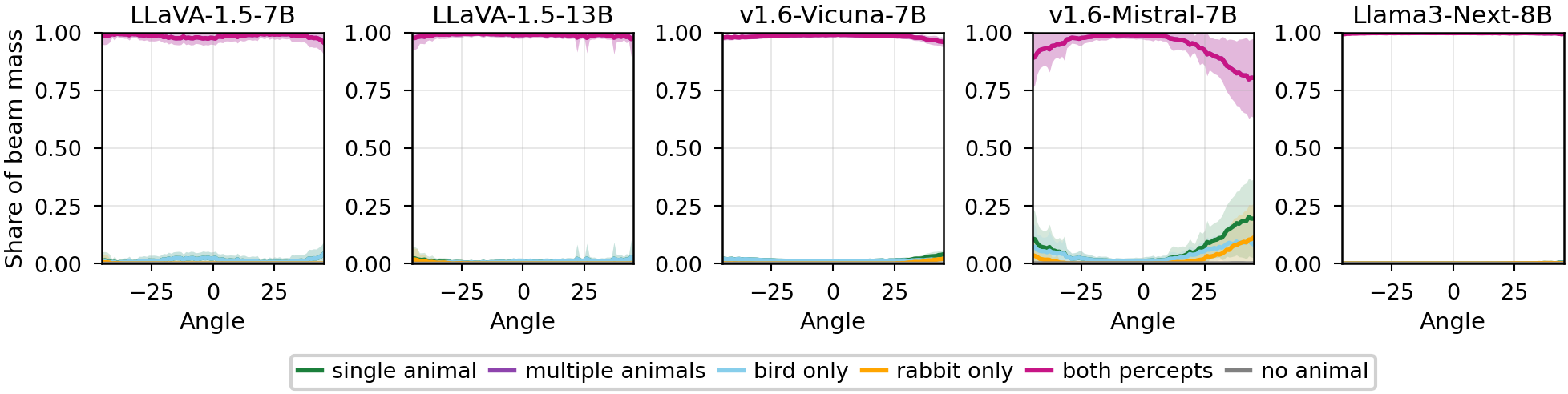}
    \caption{Beam-continuation classes on aggregated two-animal controls (default prefix) across LLaVA-family models (mean $\pm$ SD across stimuli).}
    \label{fig:appendix_control_beam}
\end{figure*}

Figure~\ref{fig:appendix_control_beam} shows the beam-continuation classes on the two-animal controls under the default prefix, over the same rotation range used for the ambiguous stimuli. Both animals are named at essentially every angle, in contrast to the ambiguous stimuli (Figure~\ref{fig:behavioral_rotation_beam_search}): the controls neither show a rotation-driven reversal of the reported interpretation nor are reported exclusively.

\subsubsection{Top-down cues on single-animal controls}
\label{sec:appendix_topdown_boundary_nonambig}

\begin{figure*}[t]
    \centering
    \includegraphics[width=\textwidth]{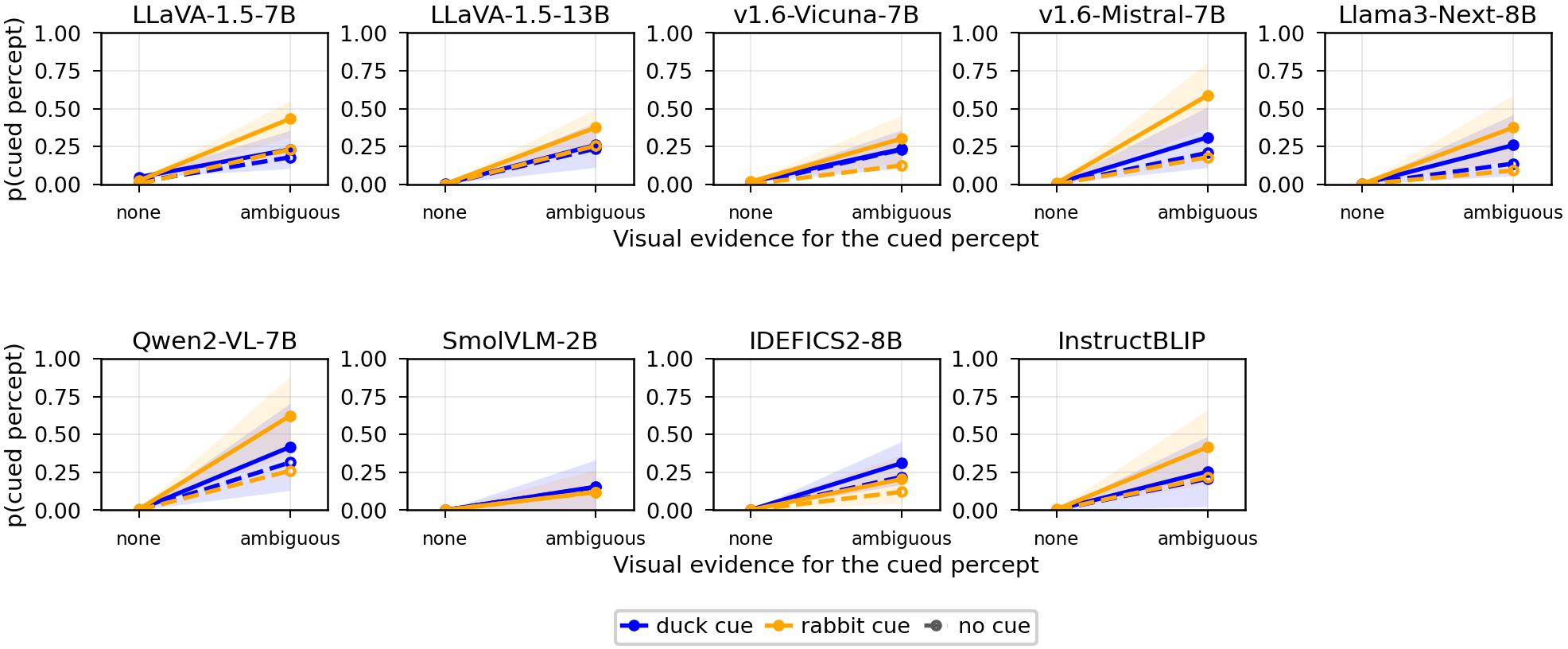}
    \caption{Prefix cues that name an interpretation (\texttt{"It starts with `d'."}, \texttt{"It starts with `r'."}) applied to stimuli offering no visual evidence for the cued percept (the image shows the other animal) or ambiguous evidence. Curves show the cued percept's probability mass with the cue (solid) and on the same stimuli with no cue (dashed), mean $\pm$ SD across stimuli, for all nine models. Without supporting evidence the cue leaves the cued percept near zero.}
    \label{fig:appendix_topdown_boundary_prefix}
\end{figure*}

\begin{figure*}[t]
    \centering
    \includegraphics[width=\textwidth]{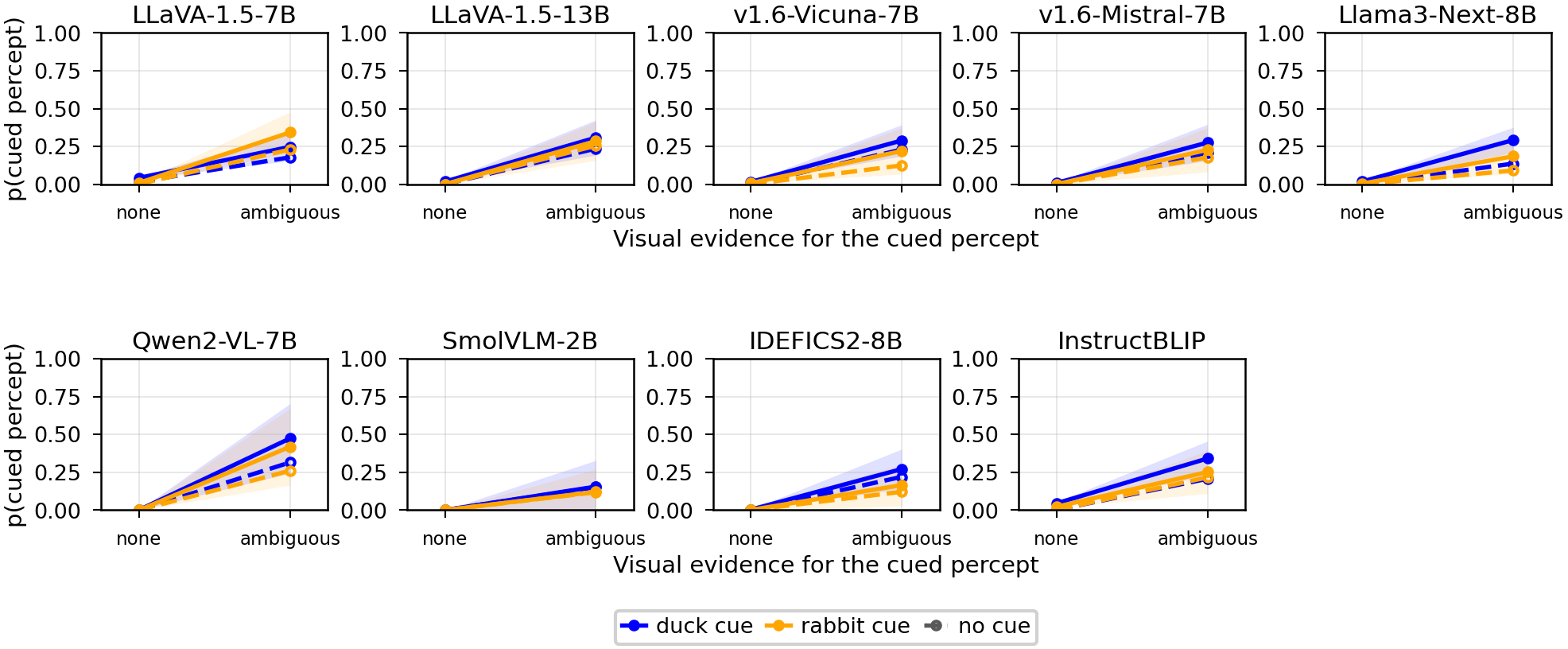}
    \caption{Semantic cues, plotted as in Figure~\ref{fig:appendix_topdown_boundary_prefix}, averaged over the four semantic cues on each side (duck: \texttt{"avian"}, \texttt{"an egg-layer"}, \texttt{"nesting"}, \texttt{"singing"}; rabbit: \texttt{"terrestrial"}, \texttt{"viviparous"}, \texttt{"herbivorous"}, \texttt{"Easter"}).}
    \label{fig:appendix_topdown_boundary_semantic}
\end{figure*}

To test whether linguistic cues can induce an interpretation unsupported by the image, we apply the same positive top-down cues to controls containing only the opposite animal. Figures~\ref{fig:appendix_topdown_boundary_prefix} and \ref{fig:appendix_topdown_boundary_semantic} compare the probability mass assigned to the cued percept when it is visually unsupported with the corresponding effect on ambiguous Visual Anagrams.

Across models, the cued percept remains near zero when the image contains only the opposite animal, whereas the same cues can substantially increase its probability on ambiguous stimuli. For example, in \textit{LLaVA-1.5-7B} the prefix cues raise the cued-percept mass to $0.23$--$0.43$ on ambiguous images but only $0.03$--$0.05$ on the corresponding single-animal controls. Thus, the top-down effect depends strongly on visual support for the cued interpretation.

The cues can nevertheless alter the remaining output distribution. On rabbit-only images, for example, the \texttt{"d"} cue in \textit{LLaVA-1.5-7B} suppresses rabbit-related output but shifts probability mainly toward other \texttt{"d"}-initial animals such as deer, dog, and donkey rather than toward duck. This suggests that linguistic constraints can influence lexical selection even when they fail to induce the visually unsupported percept.

\subsection{Generalization to figure--ground ambiguity}
\label{sec:appendix_figure_ground}

To examine whether the behavioral patterns extend beyond object-category bistability, we evaluate the Raven--Bear figure--ground stimulus used by \citet{Panagopoulou2024-xx}. We test rotation, single-letter top-down cues, beam-search exclusivity, and object count across the five LLaVA-family models. Rotation and linguistic cues continue to modulate the report, but the cue effects are more asymmetric than for duck--rabbit, and exclusivity weakens in three of the five models. Raven--Bear therefore provides a useful boundary condition rather than a uniform generalization of the duck--rabbit pattern.

Because figure--ground stimuli may more readily support a two-entity construal, we additionally compare exclusivity with the models' explicit object-count judgments. The models typically express the two candidate interpretations using bird- and dog-related tokens, which we use in the analyses below.

\subsubsection{Exclusivity under beam search}
\label{sec:appendix_figure_ground_beam}

\begin{figure*}[t]
    \centering
    \includegraphics[width=\textwidth]{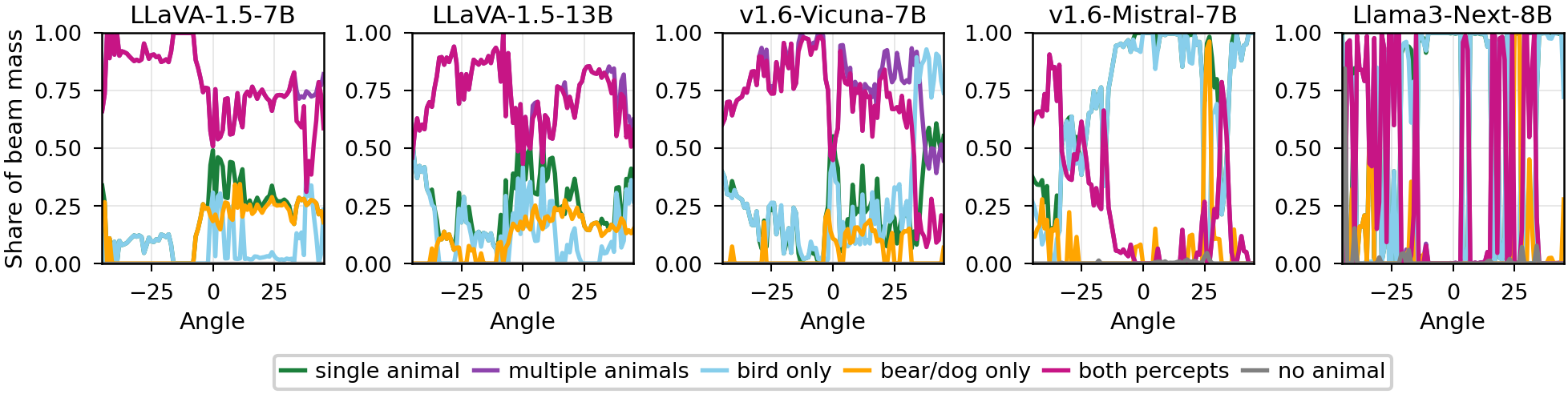}
    \caption{Raven--bear beam continuations under the default prefix, classified by the animals they name (Appendix~\ref{sec:appendix_beam_classification}), across LLaVA-family models. Here \textit{bird only} collects the bird reading (bird, raven, crow, swan), \textit{bear/dog only} the mammal reading (bear, dog, panda, polar bear), and \textit{both percepts} continuations naming one of each. \textit{Llama3-LLaVA-Next-8B} needs a longer continuation budget on this stimulus before it names an animal (Table~\ref{tab:appendix_object_count}).}
    \label{fig:appendix_figure_ground_rb_beam}
\end{figure*}

Figure~\ref{fig:appendix_figure_ground_rb_beam} shows the beam-continuation classes across rotation. At the original orientation, exclusivity is substantially weaker for \textit{LLaVA-1.5-7B}, \textit{LLaVA-1.5-13B}, and \textit{LLaVA-v1.6-Vicuna-7B}, which assign 43--51\% of beam mass to continuations naming both interpretations. In contrast, \textit{LLaVA-v1.6-Mistral-7B} and \textit{Llama3-LLaVA-Next-8B} remain fully exclusive at this orientation (Table~\ref{tab:appendix_object_count}). Rotation also changes the relative prevalence of the bird- and mammal-related reports, although the trajectories differ substantially across models.

\subsubsection{Top-down prefix cues}
\label{sec:appendix_figure_ground_topdown}

\begin{figure*}[t]
    \centering
    \includegraphics[width=\textwidth]{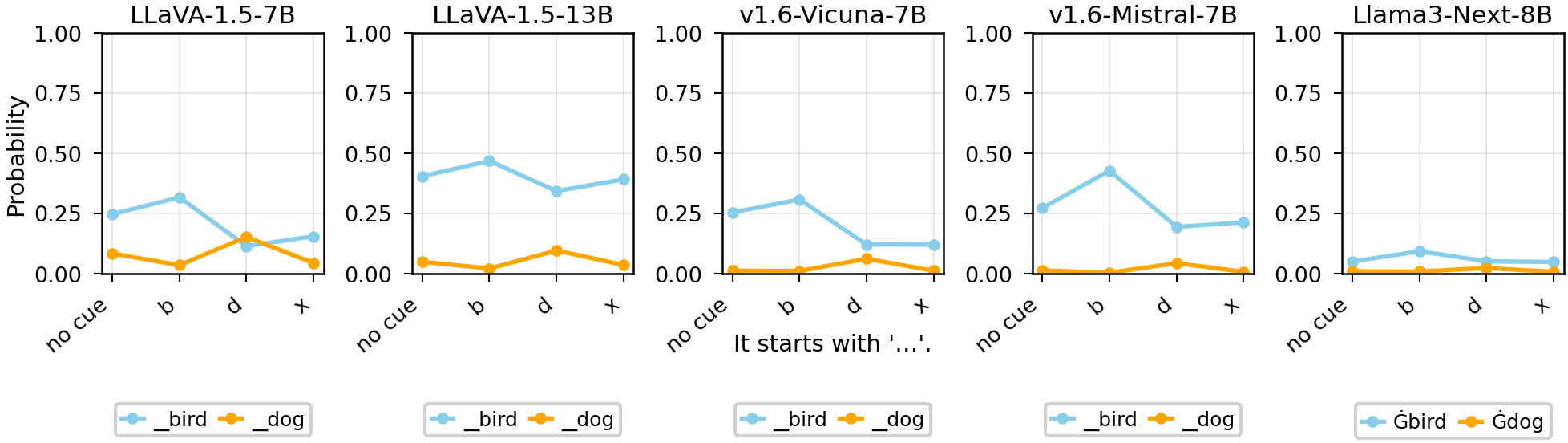}
    \caption{Raven--bear top-down single-letter prefix cues at the centre angle, across LLaVA-family models. First-token probabilities of the two percept tokens, as in the duck--rabbit top-down analyses.}
    \label{fig:appendix_figure_ground_rb_topdown}
\end{figure*}

Figure~\ref{fig:appendix_figure_ground_rb_topdown} shows the effect of single-letter prefix cues at the original orientation. The \texttt{"b"} cue increases the bird-related token in every model, whereas the \texttt{"d"} cue has a weaker and less consistent effect on the dog-related token and often acts mainly by suppressing the bird token. In \textit{LLaVA-1.5-7B}, the \texttt{"d"} cue is sufficient for the dog-related token to slightly exceed the bird-related token, but this reversal does not generalize across models. Top-down modulation therefore remains present on Raven--Bear, but is more asymmetric and less reliable than for duck--rabbit.

\subsubsection{Object count and exclusivity}
\label{sec:appendix_figure_ground_object_count}

Table~\ref{tab:appendix_object_count} compares explicit object-count judgments with beam-search exclusivity. For Harper's version and the Visual Anagrams, all five models strongly favor \texttt{"1"} and predominantly report a single animal, whereas the matched two-animal controls strongly favor \texttt{"2"} and predominantly produce multiple-animal reports.

Raven--Bear separates the models. \textit{LLaVA-1.5-7B} and \textit{LLaVA-1.5-13B} favor \texttt{"2"} and frequently name both interpretations; \textit{LLaVA-v1.6-Mistral-7B} and \textit{Llama3-LLaVA-Next-8B} favor \texttt{"1"} and remain fully exclusive; \textit{LLaVA-v1.6-Vicuna-7B} is intermediate. Thus, although the relationship is not one-to-one, exclusivity broadly covaries with the model's object-count judgment on this figure--ground stimulus.

\begin{table*}[t]
\centering
\small
\begin{tabular}{lccccc}
\toprule
Stimulus & \textit{1.5-7B} & \textit{1.5-13B} & \textit{v1.6-Vicuna} & \textit{v1.6-Mistral} & \textit{Llama3-NeXT} \\
($n$ valid VA) & (44) & (42) & (40) & (40) & (41) \\
\midrule
\multicolumn{6}{l}{\emph{Object-count query: probability of the answer \texttt{"1"} / \texttt{"2"}}} \\
\quad Harper & 0.88 / 0.10 & 0.97 / 0.03 & 0.99 / 0.00 & 0.99 / 0.01 & 0.99 / 0.01 \\
\quad Raven--Bear & 0.18 / 0.66 & 0.35 / 0.60 & 0.60 / 0.38 & 0.94 / 0.05 & 0.84 / 0.15 \\
\quad VA & 0.88 / 0.09 & 0.89 / 0.10 & 0.99 / 0.01 & 0.96 / 0.03 & 0.96 / 0.04 \\
\quad Two-animal ctrl. & 0.01 / 0.87 & 0.01 / 0.97 & 0.00 / 0.98 & 0.03 / 0.95 & 0.00 / 0.99 \\
\addlinespace
\multicolumn{6}{l}{\emph{Default query: share of beam mass naming one / multiple animals}} \\
\quad Harper & 0.93 / 0.07 & 1.00 / 0.00 & 1.00 / 0.00 & 1.00 / 0.00 & 1.00 / 0.00 \\
\quad Raven--Bear & 0.49 / 0.51 & 0.57 / 0.43 & 0.55 / 0.45 & 1.00 / 0.00 & 1.00 / 0.00 \\
\quad VA & 0.88 / 0.12 & 0.97 / 0.03 & 0.99 / 0.00 & 1.00 / 0.00 & 0.93 / 0.00 \\
\quad Two-animal ctrl. & 0.02 / 0.98 & 0.01 / 0.99 & 0.01 / 0.99 & 0.01 / 0.99 & 0.00 / 1.00 \\
\bottomrule
\end{tabular}
\caption{Object count and exclusivity across stimuli, for the five LLaVA-family models. Top block: first-token probability of the answers \texttt{"1"} and \texttt{"2"} to \texttt{"How many objects are in the image? Answer with a number only."}, read at the model's own answer step (the remainder falls on other numbers). Bottom block: shares of beam mass whose continuation names a single vs.\ multiple animals under the default query and prefix (Appendix~\ref{sec:appendix_beam_classification}; the remainder names no animal). The two blocks are separate measurements: an answer of \texttt{"1"} in the top block corresponds to exclusive reporting in the bottom one. Beam continuations use a 12-token budget, except \textit{Llama3-NeXT} on Raven--Bear, which uses 24 tokens because at 12 tokens 67\% of its beam mass had not yet named an animal. Harper and Raven--Bear at the original orientation; Visual Anagrams (VA) at each model's boundary angle and the matched two-animal controls at $0^\circ$, means over the model's valid stimuli ($n$ per column).}
\label{tab:appendix_object_count}
\end{table*}

\subsection{Robustness and boundary conditions across prompts}
\label{sec:appendix_across_prompts}

\subsubsection{Alternative task prompts}
\label{sec:appendix_alt_prompts}

We test robustness and prompt dependence using four alternative prompt framings. \textit{General description} (\texttt{"Describe this image."}) and \textit{zero-shot chain-of-thought} (\texttt{"Describe this image. Think step by step."}) test whether rotation-based modulability and exclusive reporting persist under open-ended generation. An explicit \texttt{"There are two animals."} statement tests whether linguistic context can override the default exclusive construal, while a forced-choice bird/rabbit prompt provides a separate robustness check on rotation-based modulability. All analyses use the valid Visual Anagram stimuli across all nine models; free-form generations are classified by the animals they name as in Appendix~\ref{sec:appendix_beam_classification}.

\begin{figure*}[t]
    \centering
    \includegraphics[width=\textwidth]{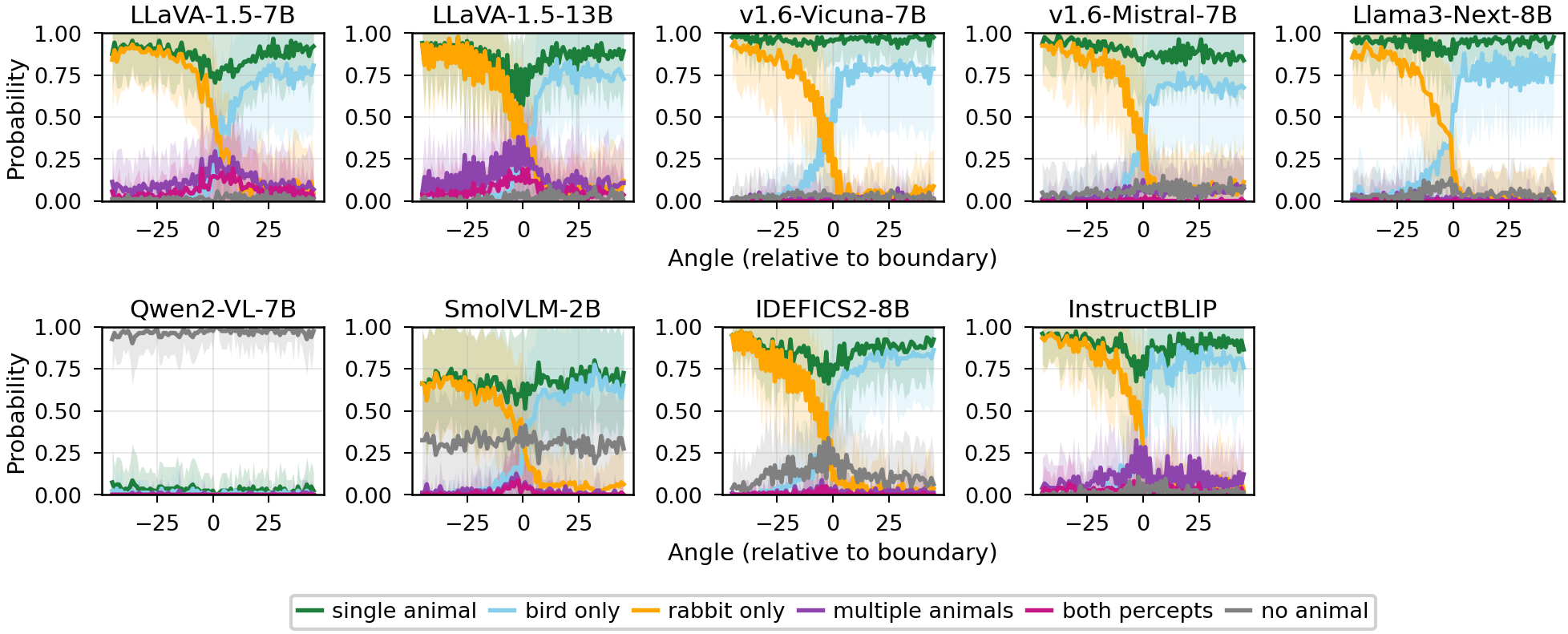}
    \caption{General description (\texttt{"Describe this image."}), averaged over the valid Visual Anagram stimuli, across all nine models. Each generation is classified by the animals it names (Appendix~\ref{sec:appendix_beam_classification}): \emph{single animal}, \emph{multiple animals} or \emph{no animal}, with \emph{bird only}/\emph{rabbit only} the subsets of single naming each percept and \emph{both percepts} the subset of multiple; 2 samples per stimulus and angle, i.e.\ 74--88 generations per angle.}
    \label{fig:appendix_across_prompts_general_va_all}
\end{figure*}

\begin{figure*}[t]
    \centering
    \includegraphics[width=\textwidth]{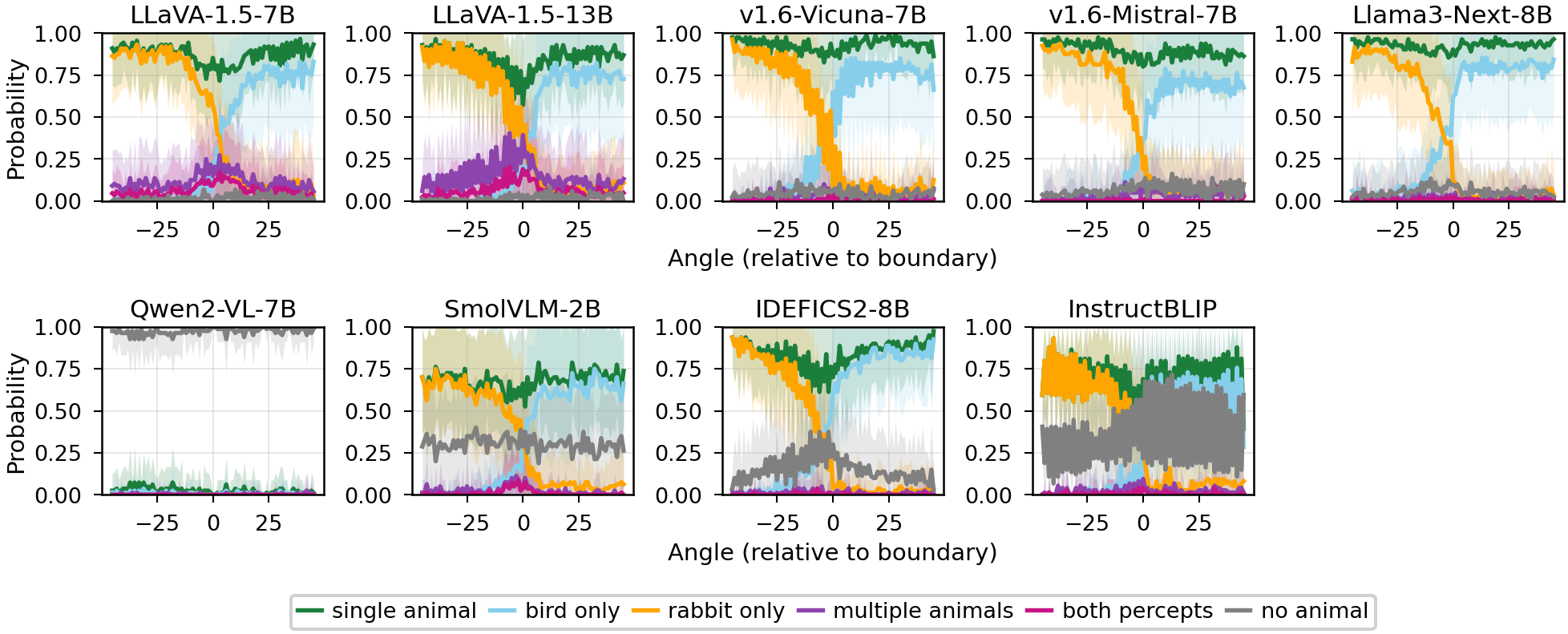}
    \caption{Zero-shot chain-of-thought, averaged over the valid Visual Anagram stimuli, across all nine models. Classes as in Figure~\ref{fig:appendix_across_prompts_general_va_all}; 2 samples per stimulus and angle, i.e.\ 74--88 generations per angle.}
    \label{fig:appendix_across_prompts_cot_va_all}
\end{figure*}

\paragraph{Open-ended descriptions.}
\label{sec:appendix_across_prompts_general}
Figures~\ref{fig:appendix_across_prompts_general_va_all} and \ref{fig:appendix_across_prompts_cot_va_all} show that both general description and zero-shot CoT preserve rotation-dependent shifts between bird- and rabbit-related reports. Single-animal descriptions also generally remain more common than multiple-animal descriptions, indicating that the qualitative exclusivity pattern persists under open-ended generation. \textit{Qwen2-VL-7B} is a notable exception in response format, producing mostly no-animal descriptions under the general-description prompt.

\begin{figure*}[t]
    \centering
    \includegraphics[width=\textwidth]{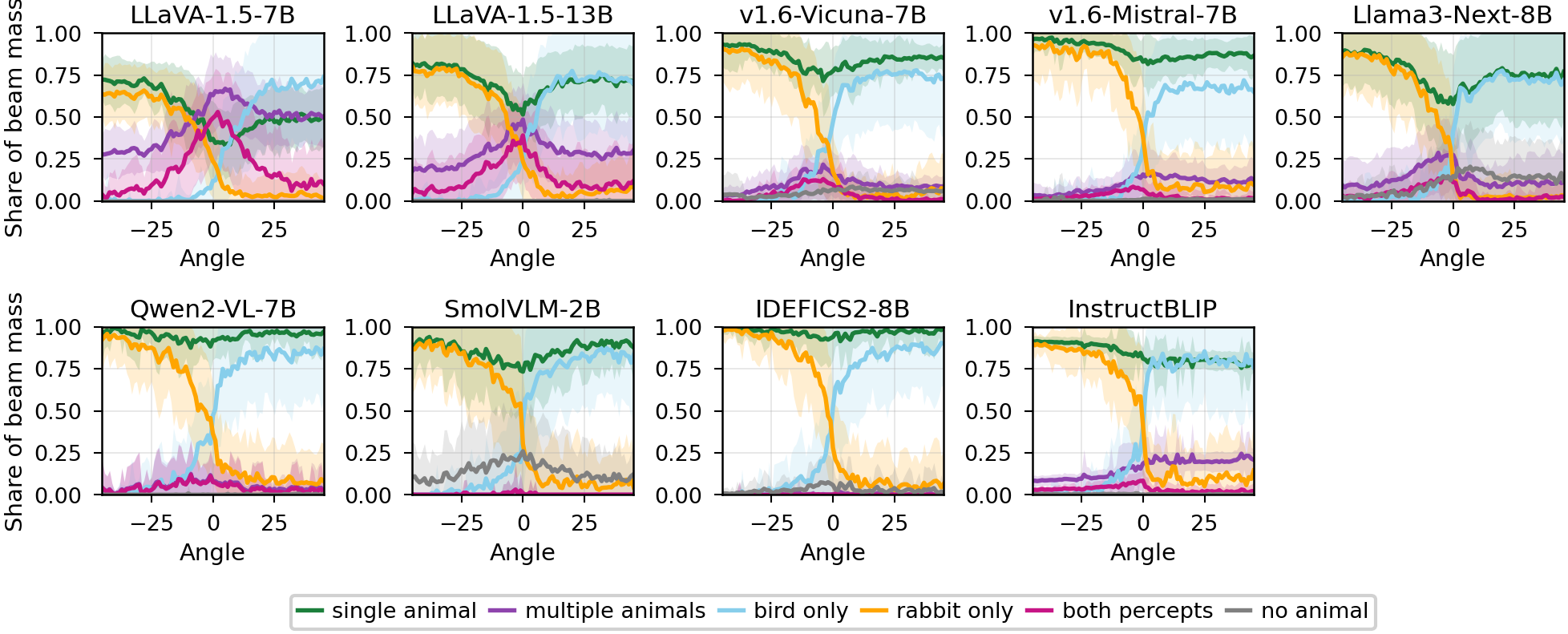}
    \caption{Beam-continuation classes on aggregated Visual Anagrams under the \texttt{"... There are two animals."} prompt (mean $\pm$ SD across stimuli).}
    \label{fig:appendix_across_prompts_two_animals_va_all}
\end{figure*}

\paragraph{Explicit two-animal hint.}
\label{sec:appendix_across_prompts_two_animals}
Figure~\ref{fig:appendix_across_prompts_two_animals_va_all} shows that explicitly stating \texttt{"There are two animals."} reduces exclusivity in a strongly model-dependent manner. Multiple-animal beam mass rises substantially for \textit{LLaVA-1.5-7B} and \textit{LLaVA-1.5-13B} (37--56\%) and more modestly for several other models (13--19\%), while \textit{Qwen2-VL-7B}, \textit{SmolVLM}, and \textit{IDEFICS2} remain largely unaffected. Thus, the default exclusive construal can be overridden to different degrees by linguistic context.

\begin{figure*}[t]
    \centering
    \includegraphics[width=\textwidth]{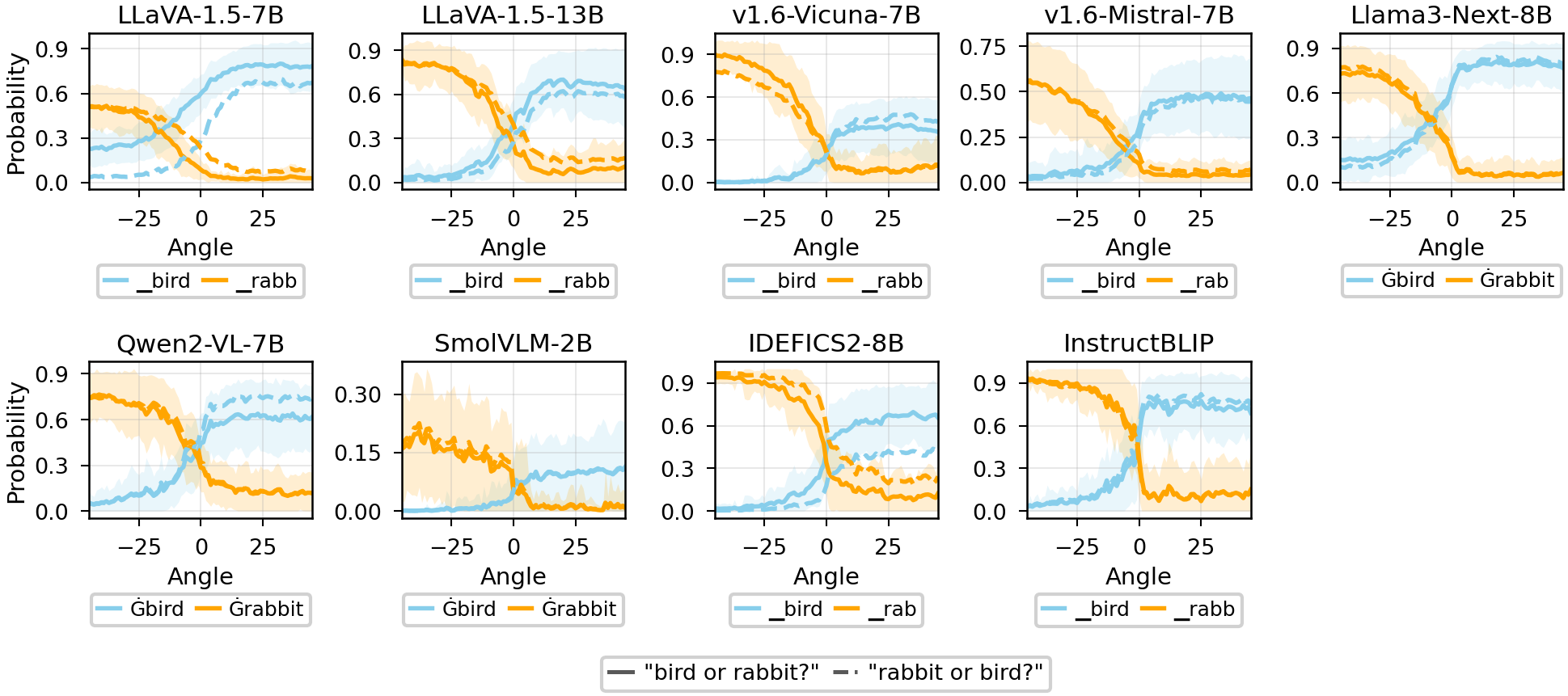}
    \caption{Forced-choice prompts (\texttt{"Is it a bird or a rabbit?"} / \texttt{"Is it a rabbit or a bird?"}, prefix \texttt{"It is a"}) on aggregated Visual Anagrams across all nine models (mean $\pm$ SD across stimuli). Solid: bird named first; dashed: rabbit named first.}
    \label{fig:appendix_forced_choice}
\end{figure*}

\paragraph{Forced choice.}
\label{sec:appendix_across_prompts_forced_choice}
Figure~\ref{fig:appendix_forced_choice} shows that all nine models retain a rotation-driven bird/rabbit reversal under both answer orders, generally near the boundary obtained with the default query. Some models nevertheless show answer-order effects, further illustrating prompt sensitivity. Because the either/or wording itself presupposes a single interpretation, we use this prompt only as a robustness check on modulability, not as a measure of exclusivity.

\subsubsection{Negative top-down cues}
\label{sec:appendix_topdown_boundary_negation}

\begin{figure*}[t]
    \centering
    \includegraphics[width=\textwidth]{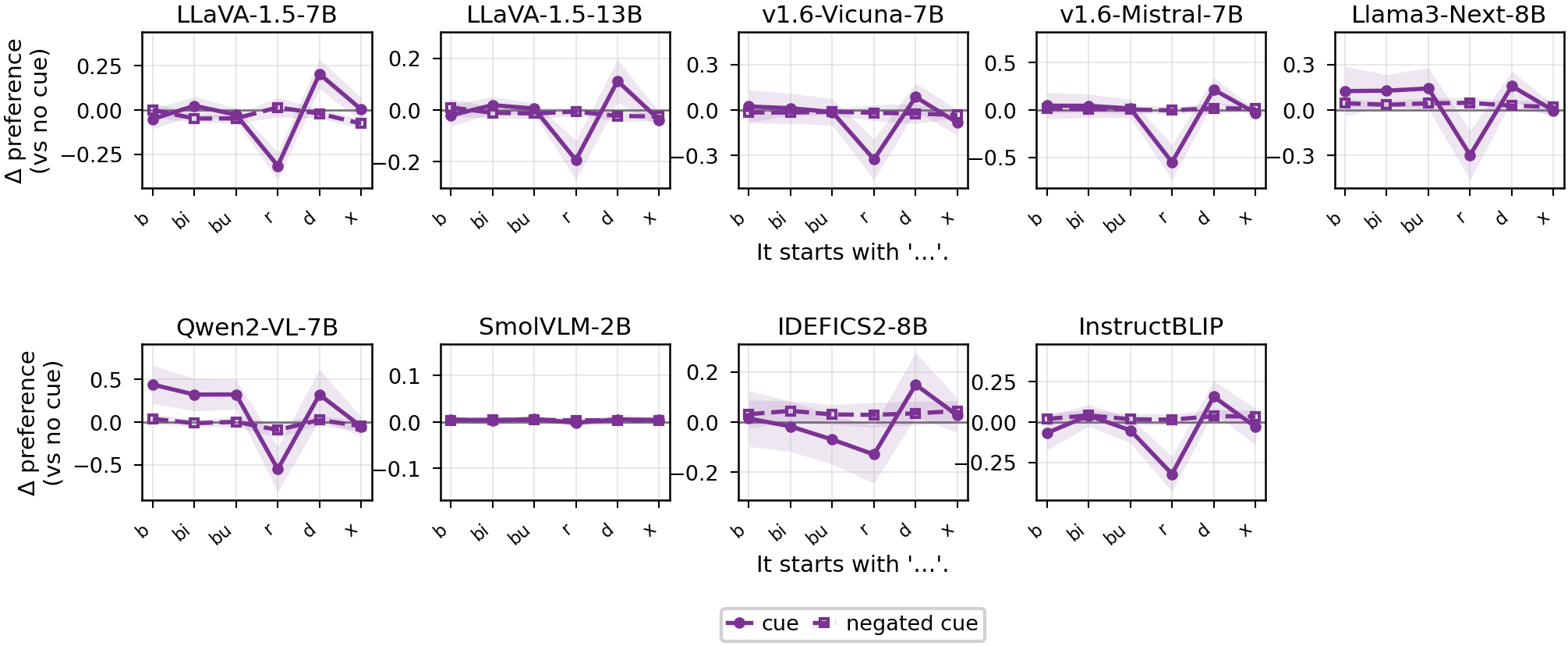}
    \caption{Prefix cues and their negations (\texttt{"It starts with `d'."} versus \texttt{"It doesn't start with `d'."}) on aggregated Visual Anagrams across all nine models. Curves show the paired shift in the duck--rabbit preference relative to the no-cue run for the same stimulus (mean $\pm$ SD across stimuli).}
    \label{fig:appendix_topdown_boundary_negation_prefix}
\end{figure*}

\begin{figure*}[t]
    \centering
    \includegraphics[width=\textwidth]{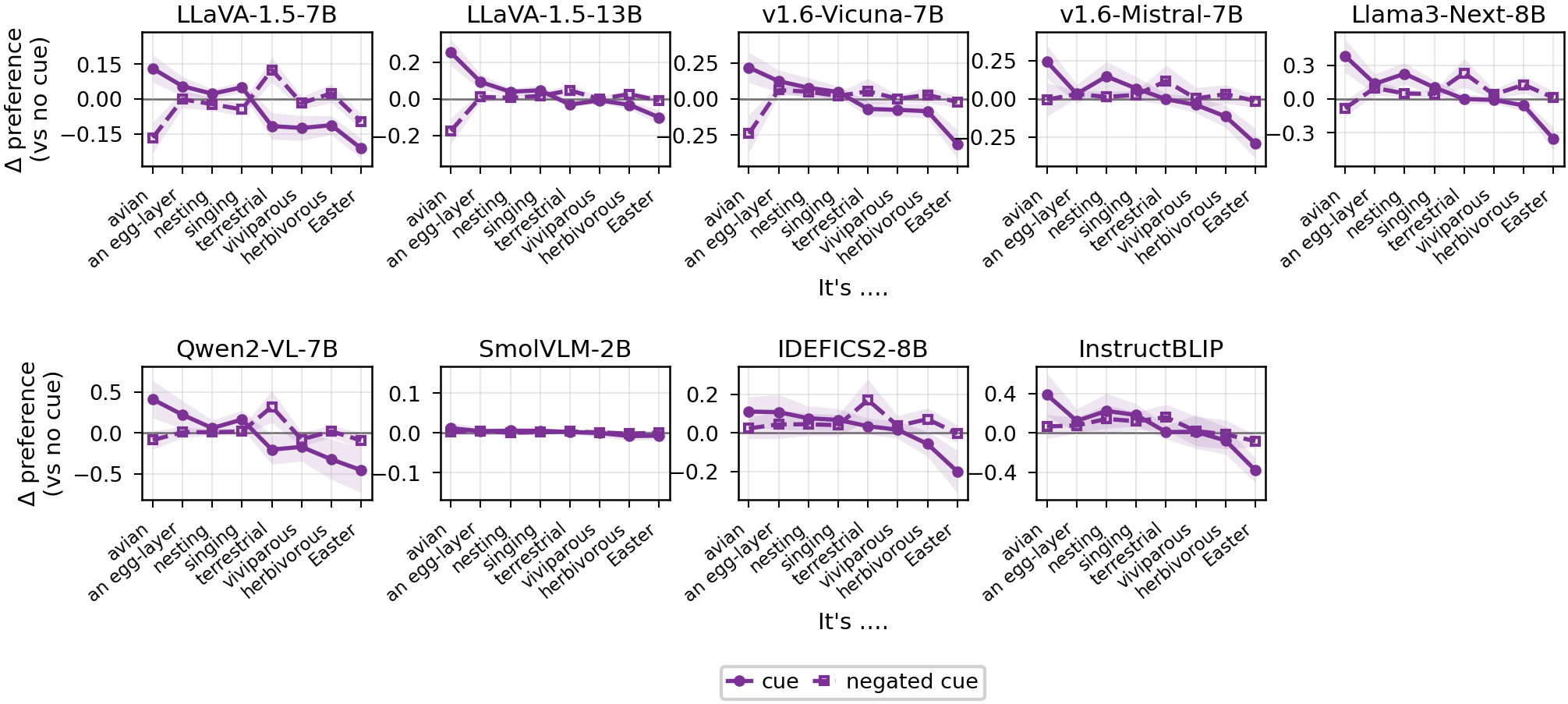}
    \caption{Semantic cues and their negations (\texttt{"It's avian."} versus \texttt{"It's not avian."}), plotted as in Figure~\ref{fig:appendix_topdown_boundary_negation_prefix}.}
    \label{fig:appendix_topdown_boundary_negation_semantic}
\end{figure*}

Figures~\ref{fig:appendix_topdown_boundary_negation_prefix} and \ref{fig:appendix_topdown_boundary_negation_semantic} test whether negating the top-down cues reverses their effects. Negated prefix cues have little effect, while semantic negation is more heterogeneous: \texttt{"not avian"} and \texttt{"not terrestrial"} often produce opposite-direction shifts, whereas most other negated semantic cues are weak and \texttt{"not Easter"} does not reverse the rabbit bias of \texttt{"Easter"}. Thus, the effect of a linguistic cue depends on how its content relates to the candidate interpretations, rather than simply on its positive or negative polarity.

\section{Additional Mechanistic Results}
\label{sec:appendix_mechanistic}

\subsection{Across LLaVA-family models}
\label{sec:appendix_mech_llava_all_models}

For space reasons, the main text reports detailed mechanistic results only for \textit{LLaVA-1.5-7B}. Here, we extend the corresponding analyses across the five LLaVA-family models where applicable. For the AnyRes models (\textit{LLaVA-v1.6-Vicuna-7B}, \textit{LLaVA-v1.6-Mistral-7B}, and \textit{Llama3-LLaVA-Next-8B}), image-token analyses operate on the 576-token global base view only; high-resolution tile tokens are left unchanged. We use the same internal-confidence threshold of 0.1 for identifying dominant image tokens in all models.

\subsubsection{Layerwise direct logit attribution}
\label{sec:appendix_mech_layerwise_dla_all_models}

\begin{figure*}[t]
    \centering
    \includegraphics[width=\textwidth]{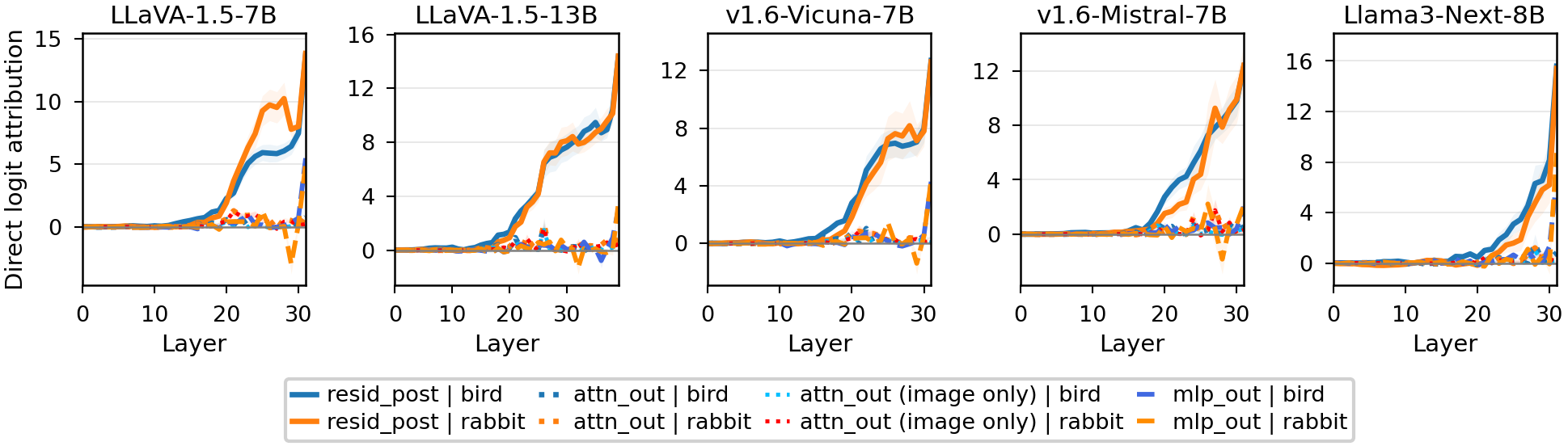}
    \caption{Layerwise direct logit attribution for the two output tokens, across all LLaVA-family models.}
    \label{fig:appendix_mech_layerwise_dla_all_models}
\end{figure*}

Figure~\ref{fig:appendix_mech_layerwise_dla_all_models} shows direct logit attribution per layer for \texttt{attn\_out}, \texttt{mlp\_out}, and \texttt{resid\_post} across all five LLaVA-family models. Across models, both interpretation logits rise primarily in the middle-to-late layers, with substantial contributions from image-token attention, although the exact onset and relative contributions of attention and MLP outputs vary across models.

\subsubsection{Bottom-up modulation: dominant patches vs.~final logit}
\label{sec:appendix_mech_bottom_up_dpm_all_models}

\begin{figure*}[t]
    \centering
    \includegraphics[width=\textwidth]{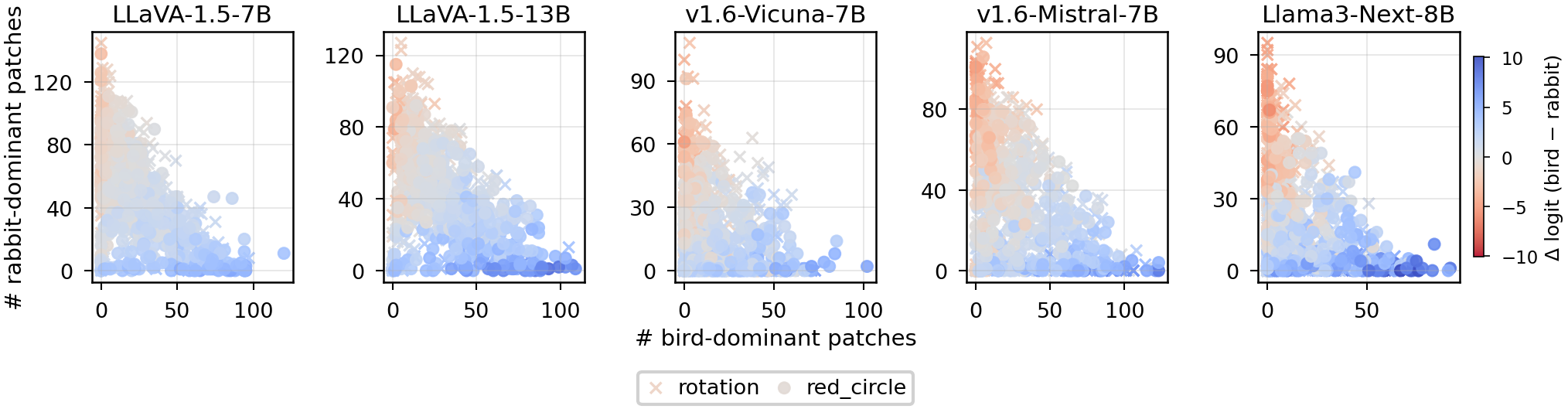}
    \caption{Bottom-up modulation: difference in number of bird/rabbit dominant patches vs.~final-token logit difference, across all LLaVA-family models.}
    \label{fig:appendix_mech_bottom_up_dpm_all_models}
\end{figure*}

Figure~\ref{fig:appendix_mech_bottom_up_dpm_all_models} extends the bottom-up analysis relating the difference in the number of bird- and rabbit-dominant image tokens to the final bird--rabbit logit difference. Across models, the balance of dominant tokens broadly tracks the final output under both rotation and red-circle modulation. This is consistent with the main-text result that bottom-up modulation is reflected in changes in image-side evidence.

\subsubsection{Top-down modulation: resampling ablation across the residual stream, attention output, and MLP output}
\label{sec:appendix_mech_top_down_patching_all_models}

\begin{figure*}[t]
    \centering
    \includegraphics[width=\textwidth]{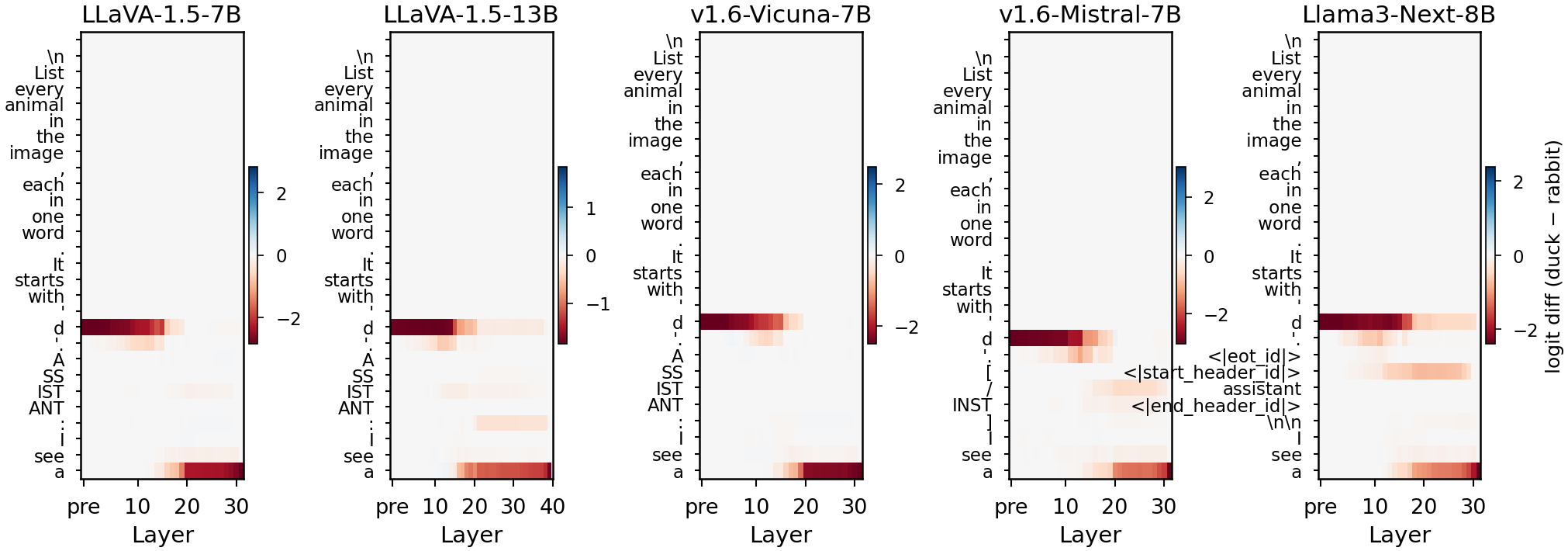}
    \caption{Top-down resampling ablation results for \texttt{resid\_post} across all LLaVA-family models, averaged over each model's valid Visual Anagram stimuli.}
    \label{fig:appendix_mech_top_down_patching_resid_post_all_models}
\end{figure*}

\begin{figure*}[t]
    \centering
    \includegraphics[width=\textwidth]{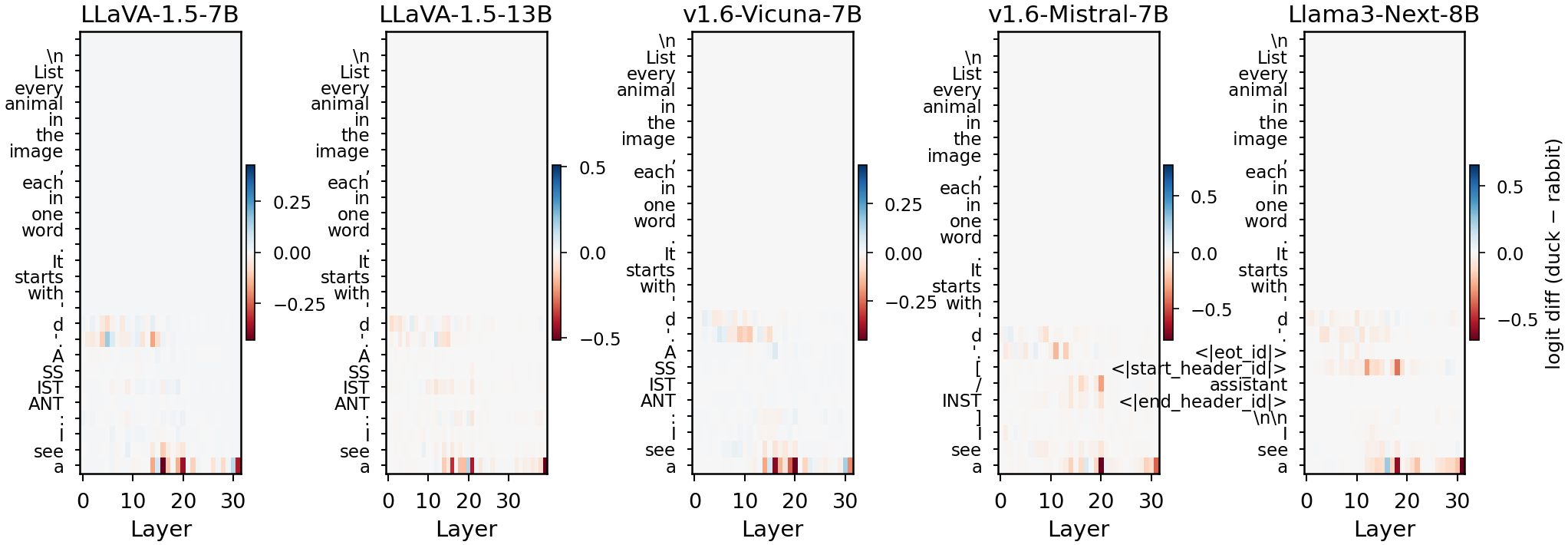}
    \caption{Top-down resampling ablation results for \texttt{attn\_out} across all LLaVA-family models, averaged over each model's valid Visual Anagram stimuli.}
    \label{fig:appendix_mech_top_down_patching_attn_out_all_models}
\end{figure*}

\begin{figure*}[t]
    \centering
    \includegraphics[width=\textwidth]{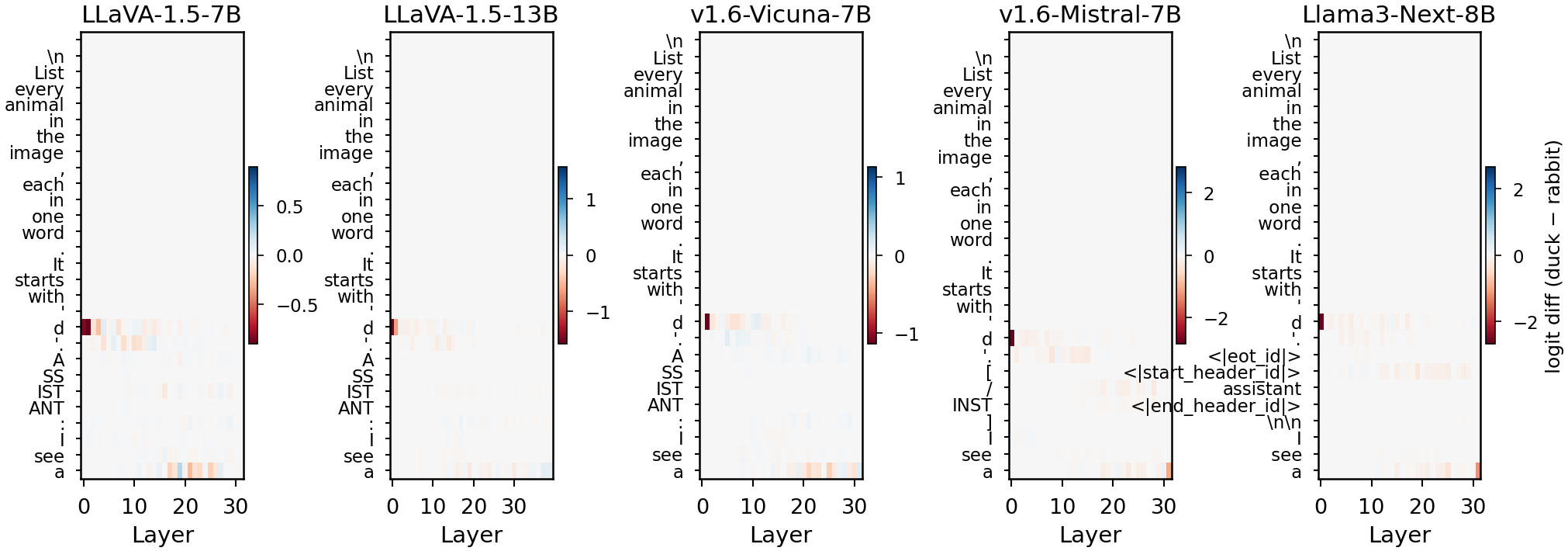}
    \caption{Top-down resampling ablation results for \texttt{mlp\_out} across all LLaVA-family models, averaged over each model's valid Visual Anagram stimuli.}
    \label{fig:appendix_mech_top_down_patching_mlp_out_all_models}
\end{figure*}

Figures~\ref{fig:appendix_mech_top_down_patching_resid_post_all_models}--\ref{fig:appendix_mech_top_down_patching_mlp_out_all_models} extend the top-down resampling-ablation analysis across all five LLaVA-family models. As in the main analysis, we patch the clean \texttt{'d'}-cue run with activations from the corrupted \texttt{'x'}-cue run for each of \texttt{resid\_post}, \texttt{attn\_out}, and \texttt{mlp\_out}. Across models, the cue effect is concentrated in middle-to-late-layer transfer from the prefix token to the final token, with substantial MLP contributions appearing earlier in the network. The precise layerwise localization and magnitude vary across models.

\subsubsection{Query patching: image-attention contribution to top-down and bottom-up modulation}
\label{sec:appendix_mech_image_attention_effect_all_models}

\begin{figure*}[t]
    \centering
    \includegraphics[width=\textwidth]{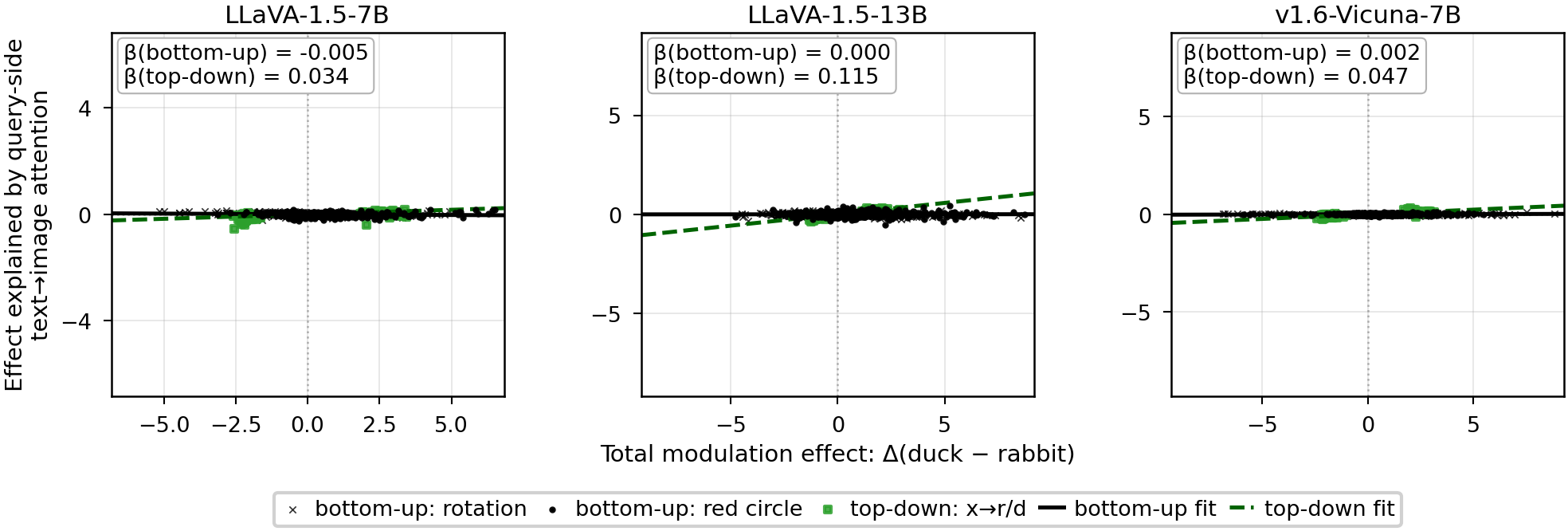}
    \caption{Query-patching results: image-attention contribution to top-down vs.~bottom-up modulation, across LLaVA-family models. \textit{LLaVA-v1.6-Mistral-7B} and \textit{Llama3-LLaVA-Next-8B} are omitted (grouped-query attention; see text).}
    \label{fig:appendix_mech_image_attention_effect_all_models}
\end{figure*}

Figure~\ref{fig:appendix_mech_image_attention_effect_all_models} extends the query-patching analysis to the three models compatible with our implementation: \textit{LLaVA-1.5-7B}, \textit{LLaVA-1.5-13B}, and \textit{LLaVA-v1.6-Vicuna-7B}. In all three models, the query-side intervention explains only a small fraction of the top-down effect and a negligible fraction of the bottom-up effect, consistent with the main-text result. We do not apply this analysis to \textit{LLaVA-v1.6-Mistral-7B} or \textit{Llama3-LLaVA-Next-8B} because their grouped-query attention is incompatible with our implementation, which assumes matched query and key/value heads.

\subsubsection{Object-count patching and exclusivity}
\label{sec:appendix_mech_exclusivity_all_models}

\begin{figure*}[t]
    \centering
    \includegraphics[width=\textwidth]{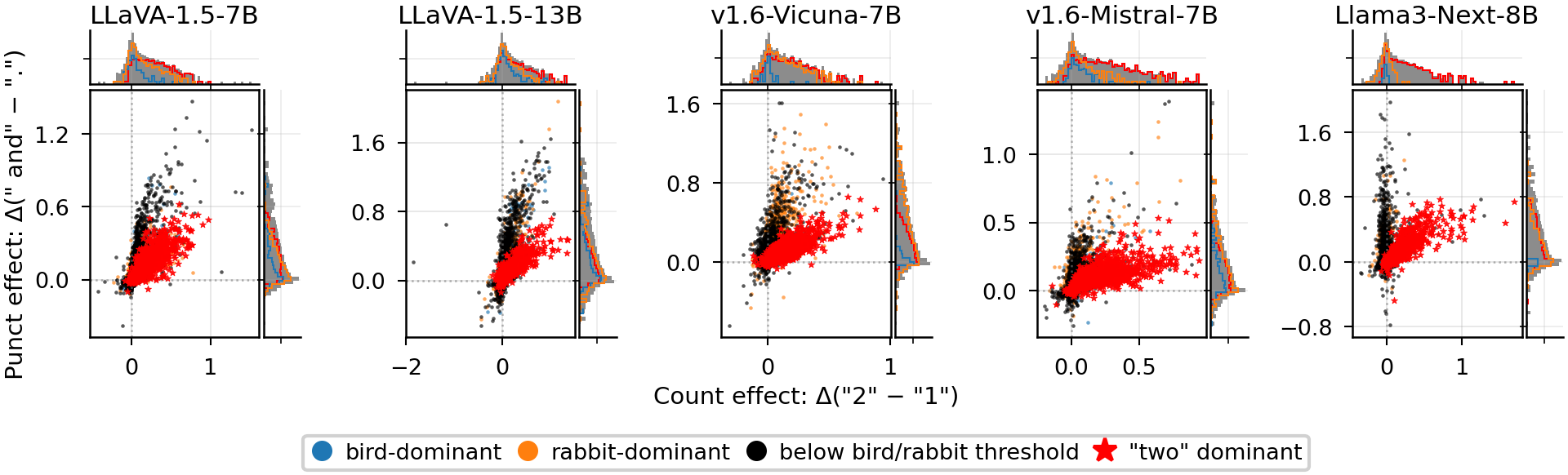}
    \caption{Exclusivity probe (object-count vs.~continuation effect of single-token image-embedding patching) across all LLaVA-family models.}
    \label{fig:appendix_mech_exclusivity_all_models}
\end{figure*}

Figure~\ref{fig:appendix_mech_exclusivity_all_models} extends the object-count patching analysis across all five LLaVA-family models. Across models, image-token interventions that increase the explicit \texttt{"2"}--\texttt{"1"} object-count contrast also tend to increase the \texttt{"and"}--\texttt{"."} continuation contrast, supporting the association between object-count processing and exclusive reporting observed in \textit{LLaVA-1.5-7B}.

Direct decoding of \texttt{"2"} is not by itself a marker of causal relevance: although some \texttt{"2"}-dominant image tokens have substantial effects on the count contrast, most have little individual effect on either measure, consistent with prior observations that number-like logit-lens readouts at image positions need not correspond to causally relevant visual features \citep{Neo2024-ep,Neo2026-fw}. \textit{Llama3-LLaVA-Next-8B} additionally shows a set of image tokens, mostly without a dominant readout, whose interventions substantially affect the \texttt{"and"}--\texttt{"."} contrast while having little effect on the explicit \texttt{"2"}--\texttt{"1"} contrast. Thus, the relationship between object-count processing and exclusive reporting generalizes across models, while the \textit{Llama3-LLaVA-Next-8B} result indicates that object-count encoding is not the sole factor determining exclusivity.

\section{Additional Discussion and Future Directions}
\label{sec:additional_discussion}

The main text focuses on the limitations most directly relevant to interpreting our present claims. Here, we expand on several broader directions that would extend the framework beyond the current static, object-category setting, including switching dynamics and active visual sampling, history dependence across repeated presentations, other forms of bistable perception, and relational reinterpretation across multiple ambiguous figures.

\paragraph{Switching dynamics and active visual sampling.}
A central difference between our setup and human bistable perception is that the present experiments do not capture spontaneous alternations between interpretations during continued viewing. In humans, a fixed ambiguous stimulus can alternate between perceptual interpretations over time, and this process is shaped by eye movements, adaptation, and selective fixation \citep{Brascamp2018-gq,Hsu2025-sj}. By contrast, the MLLMs studied here encode a fixed image and produce a linguistic report. Report-level changes can be elicited by asking a model to ``look again,'' prompting longer reasoning with an appended ``wait'' \citep{Muennighoff2025-ki}, or increasing sampling temperature, but such changes are better interpreted as re-evaluation or stochastic decoding from a fixed visual representation than as human-like perceptual switching under ongoing visual input.

A closer model-side analogue would require active-perception mechanisms that acquire new visual evidence over time, for example through shifting fixation, cropping, or zooming. Active-inference accounts provide a theoretically grounded, brain-oriented explanation of perceptual alternations in terms of overt or covert sampling actions that reduce uncertainty \citep{Parr2019-bp,Novicky2024-co}; existing active-inference models of bistable perception make this account tractable by specifying the relevant perceptual hypotheses and sampling variables within a phenomenon-specific generative model. A complementary question is which internal variables (e.g., object count) are causally relevant to different aspects of bistable perception in a broadly learned visual system when the competing percepts are not explicitly built into the model. Active-perception MLLMs such as DeepEyes \citep{Zheng2025-hx} provide a complementary testbed: whereas local cueing is imposed externally in our red-circle manipulation, these models can select new visual evidence themselves, allowing us to ask whether switching-like dynamics and their underlying mechanisms can emerge in a general-purpose model. Findings from such systems could in turn help constrain the structure of richer generative models of active perception.

\paragraph{History dependence across repeated presentations.}
A separate limitation is that our experiments treat each image--prompt pair independently, whereas human multistable perception is history-dependent: the interpretation reported on a given presentation can depend on previous perceptual states, including stabilization effects after intermittent viewing \citep{Pastukhov2008-eq}. A natural extension would be to compare reports across clockwise and counterclockwise sequences of the same rotated duck--rabbit images. Such an experiment would extend modulability from static stimulus--response effects to hysteresis-like dependence on prior perceptual states.

\paragraph{Other forms of bistable perception.}
The present study focuses primarily on static object-category bistability, where a single image region supports incompatible object identities. Bistable and multistable perception also includes other stimulus classes, such as geometric or depth ambiguity, figure--ground ambiguity, motion-based ambiguity, and binocular rivalry \citep{Leopold1999-ae,Brascamp2018-gq,Meng2004-zh}, which require different operationalizations. Necker-cube-style depth ambiguity \citep{Necker1832-jj}, for example, involves viewpoint or relational interpretations rather than object-category labels, making clean token-level measures difficult to define. Motion-based ambiguity, such as the motion quartet \citep{Ramachandran1986-dw}, depends on temporal evidence and is therefore not well captured by the static image-input models studied here. Figure--ground stimuli such as Rubin's vase \citep{Rubin1915-lg} are closer to our setting because they are also single-image ambiguous figures. However, unlike duck--rabbit-like object-category ambiguities, for which Visual Anagrams provide novel synthetic variants, it is less straightforward to generate synthetic figure--ground stimuli that preserve the relevant ambiguity while controlling for memorization. Our Raven--Bear analysis provides an initial boundary condition for this stimulus class (Section~\ref{sec:behavioral_exclusivity}; Appendix~\ref{sec:appendix_figure_ground}), but systematic generalization across figure--ground and other forms of bistability remains for future work.

\paragraph{Relational reinterpretation of multiple ambiguous figures.}
Our operationalization of exclusivity focuses on whether a single ambiguous object is reported as one interpretation rather than two. It does not address cases in which multiple identical ambiguous figures can be organized into a relational scene. For example, relational cues such as ``duck eats rabbit'' can help human observers see two identical duck--rabbit figures as playing different roles, even though many observers cannot initially see them as different interpretations side by side \citep{Jensen2011-qj,Mathewson2018-of}. This phenomenon is not simply a failure of exclusivity for a single object: the two interpretations are assigned to different instances and structured by a higher-level relational context. A natural extension of our framework would be to present two identical ambiguous figures with relational priors and ask questions such as \texttt{"What tries to eat what?"}. This would test whether linguistic context can induce a coherent relational construal over multiple ambiguous objects, a form of top-down modulation not addressed here.

\end{document}